\documentclass[]{fairmeta}

\usepackage{amsmath,amsfonts,bm}

\usepackage{xspace}
\newcommand*{\eg}{{\it e.g.}\@\xspace}
\newcommand*{\ie}{{\it i.e.}\@\xspace}

\def\eqref#1{equation~\ref{#1}}

\def\1{\bm{1}}

\DeclareMathAlphabet{\mathsfit}{\encodingdefault}{\sfdefault}{m}{sl}
\SetMathAlphabet{\mathsfit}{bold}{\encodingdefault}{\sfdefault}{bx}{n}

\def\gT{{\mathcal{T}}}

\def\gX{{\mathcal{X}}}

\newcommand{\E}{\mathbb{E}}

\newcommand{\R}{\mathbb{R}}

\usepackage{hyperref}
\usepackage{url}
\usepackage{listings}
\usepackage{graphicx}
\usepackage{array}
\usepackage{booktabs}
\usepackage{multirow}
\usepackage{array}
\usepackage{graphicx}
\usepackage{wrapfig} 
\usepackage{color,soul}
\usepackage{subcaption}

\usepackage{algorithm}
\usepackage{algpseudocode}
\usepackage{eqnarray}
\usepackage{csquotes}

\usepackage{sidecap}

\renewcommand{\eqref}[1]{(\ref{#1})}

\newcommand{\image}{\texttt{<|image|>}}

\newcommand{\strip}{f_{\text{rm-blanks}}}
\newcommand{\blank}{{\color{gray}\varepsilon}}

\colorlet{mygreen}{green!55!black}
\definecolor{myyellow1}{HTML}{D89A3C}
\colorlet{myyellow}{myyellow1!90!black}
\definecolor{metablue}{HTML}{0064E0}

\definecolor{cadetblue}{rgb}{0.37, 0.62, 0.63}

\newcommand{\hlb}[1]{\colorbox{metablue!25}{#1}}
\newcommand{\hlp}[1]{\colorbox{pink!50}{#1}}

\newcommand{\cellhi}{\cellcolor{metablue!10}}
\newcommand{\cellhii}{\cellcolor{metablue!25}}

\newcommand{\cellhjj}{\cellcolor{myyellow!30}}

\usepackage{xcolor} 
\definecolor{burgundy}{HTML}{800020}

\newcommand{\AR}{AR\xspace}
\newcommand{\EF}{OneFlow\xspace} 

\usepackage[table]{xcolor}
\usepackage{booktabs}
\usepackage{adjustbox}
\usepackage{tabularx} 
\usepackage{rotating}
\definecolor{colGen}{rgb}{0.92,0.97,0.98}
\definecolor{colKnow}{rgb}{0.89,0.95,0.97}
\definecolor{colOCR}{rgb}{0.85,0.93,0.93}
\definecolor{colVision}{rgb}{0.83,0.90,0.95}
\definecolor{colHalluc}{rgb}{0.93,0.91,0.98}

\renewcommand{\P}{\mathbb{P}}
\definecolor{userbg}{RGB}{245, 245, 245}
\definecolor{userborder}{RGB}{210, 229, 255}
\definecolor{userfont}{RGB}{0, 0, 0}
\definecolor{EditTeal}{RGB}{5, 146, 148}
\definecolor{EditDarkRed}{RGB}{137, 30, 83}
\definecolor{EditOrange}{RGB}{204, 138, 47}
\usepackage{pgf}
\usepackage{adjustbox}
\usepackage{tcolorbox}
\tcbuselibrary{most}

\newcommand{\ins}{{{\text{ins}}}}

\newcommand{\tsum}{\textstyle \sum}

\definecolor{citecolor}{HTML}{0071bc}
\hypersetup{
    colorlinks=true,%
    citecolor=citecolor,%
    filecolor=citecolor,%
    linkcolor=citecolor,%
    urlcolor=citecolor
}
\usepackage{multirow}
\usepackage{colortbl} 
\usepackage{booktabs}
\usepackage{tcolorbox} 
\usepackage{enumitem}
\usepackage{graphicx}
\usepackage{multirow}
\usepackage{array}
\definecolor{listcolor}{RGB}{50,120,230}
\usepackage{arydshln}
\usepackage{url}
\usepackage{caption}
\usepackage{xcolor}
\usepackage{makecell}

\newcounter{researchquestion}

\newcommand{\researchquestion}[2][]{%
  \vspace{0.8em}
  \refstepcounter{researchquestion}
  \begin{tcolorbox}[
    enhanced,
    colback=blue!5,
    colframe=listcolor,
    fonttitle={\fontsize{10.5pt}{12.8pt}\selectfont\bfseries\color{blue!20!black}},  
    title=Question \theresearchquestion,
    toprule=1.5pt,
    bottomrule=0.8pt,
    leftrule=0.8pt,
    rightrule=0.8pt,
    left=6pt,
    right=6pt,
    top=6pt,
    bottom=6pt,
    boxsep=3pt
  ]
  \normalsize #2
  \end{tcolorbox}
  \ifx\\#1\\\else\label{rq:#1}\fi
  \vspace{0.5em}
}

\title{OneFlow: Concurrent Mixed-Modal and \\Interleaved Generation with Edit Flows}

\author[1]{John Nguyen}
\author[1]{Marton Havasi}
\author[1,2]{Tariq Berrada}
\author[1]{Luke Zettlemoyer}
\author[1]{Ricky T. Q. Chen}
\affiliation[1]{FAIR at Meta}
\affiliation[2]{Univ. Grenoble Alpes, Inria, CNRS, Grenoble INP, LJK, France}

\abstract{We present \textbf{\EF}, the first non-autoregressive multimodal model that enables variable-length and concurrent mixed-modal generation. Unlike autoregressive models that enforce rigid causal ordering between text and image generation, \EF combines an insertion-based Edit Flow for discrete text tokens with Flow Matching for image latents. \EF enables concurrent text-image synthesis with hierarchical sampling that prioritizes content over grammar. 
Through controlled experiments across model sizes from 1B to 8B, we demonstrate that \EF outperforms autoregressive baselines on both generation and understanding tasks while using up to 50\% fewer training FLOPs. \EF surpasses both autoregressive and diffusion-based approaches while unlocking new capabilities for concurrent generation, iterative refinement, and natural reasoning-like generation.}

\metadata[Website]{\href{https://oneflow.framer.ai/}{https://oneflow.framer.ai/}}

\begin{document}

\maketitle

\begin{figure}[!htb]
    \centering
    \includegraphics[width=\linewidth]{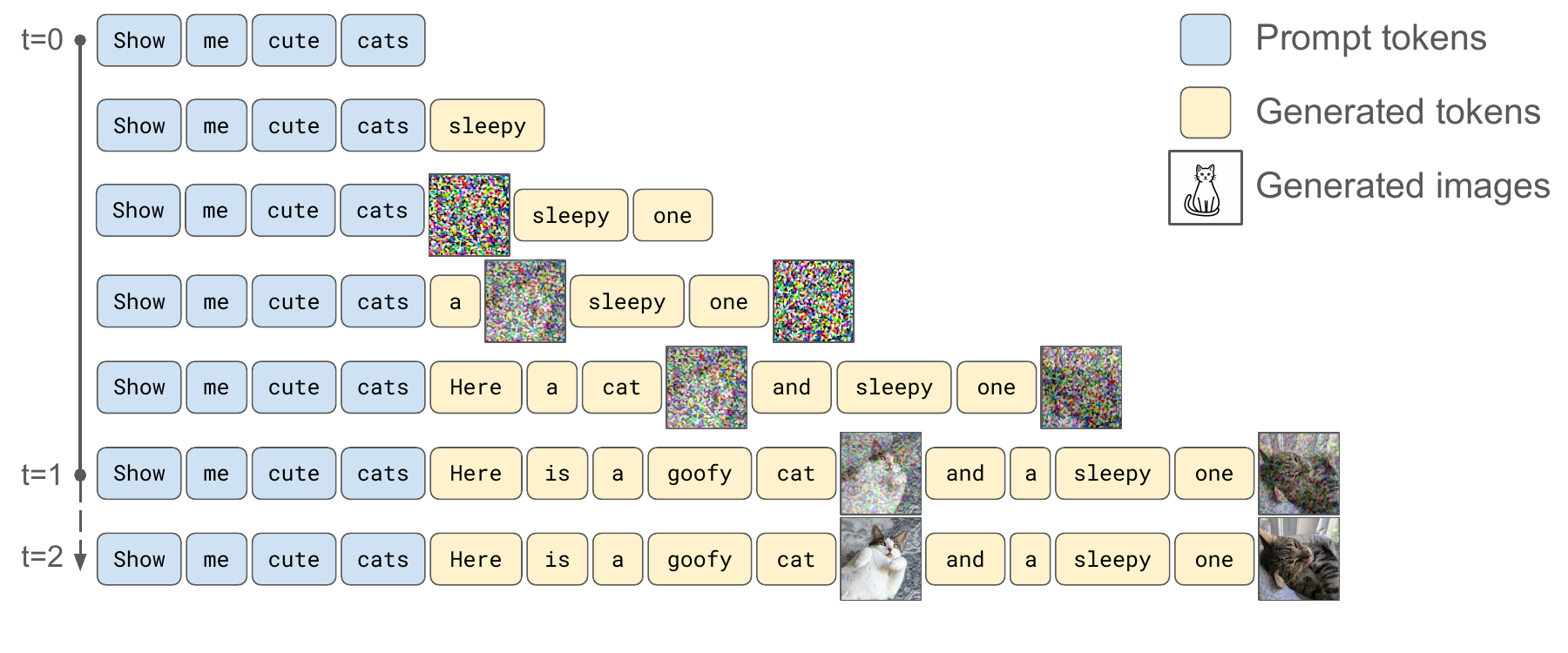}\vspace{-1.5em}
    \caption{OneFlow is a variable-length non-autoregressive model that can concurrently generate interleaved text and variable number of images using insertions as a primitive operation.}
    \label{fig:joint_generation}
\end{figure}

\begin{figure}
    \centering
    \includegraphics[width=\linewidth, trim = 0px 42px 0px 0px, clip]{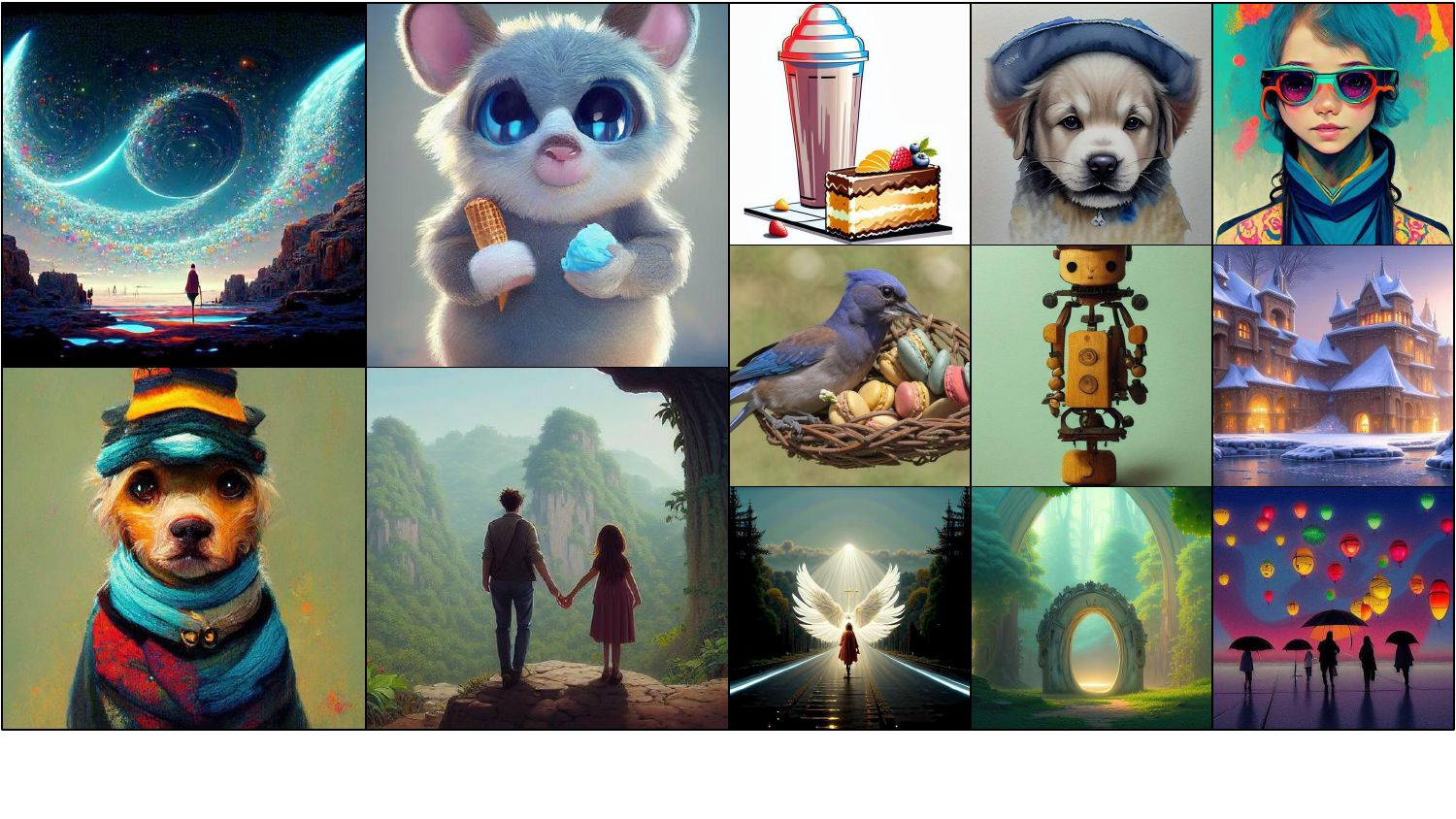}
    \caption{\textbf{Text-to-image generation.} Generated images at 512$\times$512 resolution from OneFlow. Prompts are in Figure \ref{fig:generated_images_with_prompts}.}
    \label{fig:generated_images_collage}
\end{figure}
\begin{figure}
    \centering
    \begin{minipage}{\linewidth}
        \centering
        \includegraphics[width=\linewidth,height=\textheight,keepaspectratio]{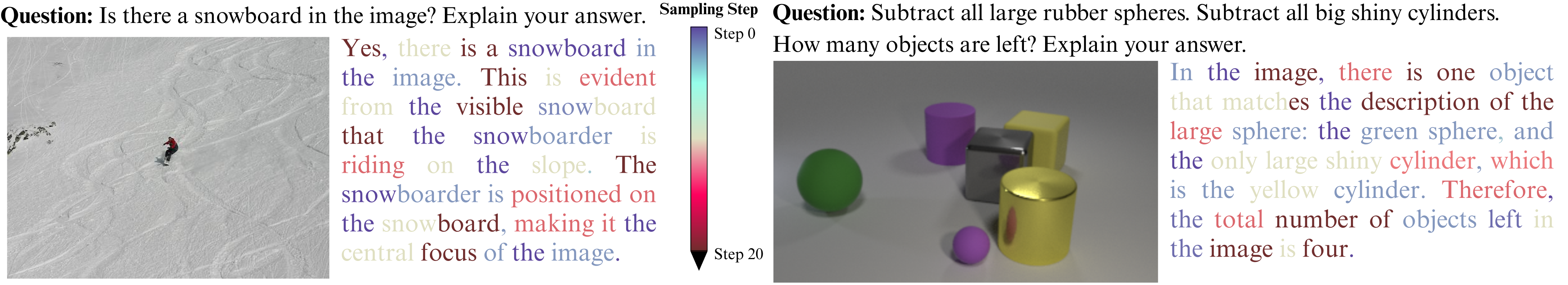}
        \caption{\textbf{Visual question answering.} \EF{} generates by simply inserting tokens based on confidence, resulting in a natural hierarchical sampling and implicit reasoning where the most difficult answer tokens are generated later.}
        \label{fig:vqa_viz}

    \end{minipage}
\end{figure}
\begin{figure}
\resizebox{\textwidth}{!}{
\begin{tabular}{m{0.9\linewidth} m{0.18\linewidth}}
\toprule
\multicolumn{2}{l}{
Generated tokens: \hfill $t=0$ \colorbox[HTML]{000000}{\rule{-0.2em}{0.7em}}\colorbox[HTML]{060103}{\rule{-0.2em}{0.7em}}\colorbox[HTML]{0d0306}{\rule{-0.2em}{0.7em}}\colorbox[HTML]{14050a}{\rule{-0.2em}{0.7em}}\colorbox[HTML]{1b060d}{\rule{-0.2em}{0.7em}}\colorbox[HTML]{220811}{\rule{-0.2em}{0.7em}}\colorbox[HTML]{290a14}{\rule{-0.2em}{0.7em}}\colorbox[HTML]{300c18}{\rule{-0.2em}{0.7em}}\colorbox[HTML]{370d1b}{\rule{-0.2em}{0.7em}}\colorbox[HTML]{3e0f1f}{\rule{-0.2em}{0.7em}}\colorbox[HTML]{451022}{\rule{-0.2em}{0.7em}}\colorbox[HTML]{4a0e26}{\rule{-0.2em}{0.7em}}\colorbox[HTML]{4f0d29}{\rule{-0.2em}{0.7em}}\colorbox[HTML]{540b2d}{\rule{-0.2em}{0.7em}}\colorbox[HTML]{590930}{\rule{-0.2em}{0.7em}}\colorbox[HTML]{5f0734}{\rule{-0.2em}{0.7em}}\colorbox[HTML]{640637}{\rule{-0.2em}{0.7em}}\colorbox[HTML]{69043a}{\rule{-0.2em}{0.7em}}\colorbox[HTML]{6e023e}{\rule{-0.2em}{0.7em}}\colorbox[HTML]{730141}{\rule{-0.2em}{0.7em}}\colorbox[HTML]{790446}{\rule{-0.2em}{0.7em}}\colorbox[HTML]{7e0e4b}{\rule{-0.2em}{0.7em}}\colorbox[HTML]{831850}{\rule{-0.2em}{0.7em}}\colorbox[HTML]{882355}{\rule{-0.2em}{0.7em}}\colorbox[HTML]{8d2d5a}{\rule{-0.2em}{0.7em}}\colorbox[HTML]{933860}{\rule{-0.2em}{0.7em}}\colorbox[HTML]{984265}{\rule{-0.2em}{0.7em}}\colorbox[HTML]{9d4d6a}{\rule{-0.2em}{0.7em}}\colorbox[HTML]{a2576f}{\rule{-0.2em}{0.7em}}\colorbox[HTML]{a76174}{\rule{-0.2em}{0.7em}}\colorbox[HTML]{ac687a}{\rule{-0.2em}{0.7em}}\colorbox[HTML]{af6b7f}{\rule{-0.2em}{0.7em}}\colorbox[HTML]{b36f84}{\rule{-0.2em}{0.7em}}\colorbox[HTML]{b67289}{\rule{-0.2em}{0.7em}}\colorbox[HTML]{b9758e}{\rule{-0.2em}{0.7em}}\colorbox[HTML]{bd7994}{\rule{-0.2em}{0.7em}}\colorbox[HTML]{c07c99}{\rule{-0.2em}{0.7em}}\colorbox[HTML]{c4809e}{\rule{-0.2em}{0.7em}}\colorbox[HTML]{c783a3}{\rule{-0.2em}{0.7em}}\colorbox[HTML]{cb87a8}{\rule{-0.2em}{0.7em}}\colorbox[HTML]{ce8dae}{\rule{-0.2em}{0.7em}}\colorbox[HTML]{d294b3}{\rule{-0.2em}{0.7em}}\colorbox[HTML]{d59bb8}{\rule{-0.2em}{0.7em}}\colorbox[HTML]{d9a2bd}{\rule{-0.2em}{0.7em}}\colorbox[HTML]{dca9c2}{\rule{-0.2em}{0.7em}}\colorbox[HTML]{e0b0c8}{\rule{-0.2em}{0.7em}}\colorbox[HTML]{e3b7cd}{\rule{-0.2em}{0.7em}}\colorbox[HTML]{e7bed2}{\rule{-0.2em}{0.7em}}\colorbox[HTML]{eac5d7}{\rule{-0.2em}{0.7em}}\colorbox[HTML]{EECCDD}{\rule{-0.2em}{0.7em}}
 $t=1$
} \\
\midrule
{\color[HTML]{000000}\textbf{I}} {\color[HTML]{000000}\textbf{want}} {\color[HTML]{000000}\textbf{to}} {\color[HTML]{000000}\textbf{paint}} {\color[HTML]{000000}\textbf{flowers}} {\color[HTML]{000000}\textbf{but}} {\color[HTML]{000000}\textbf{I}} {\color[HTML]{000000}\textbf{need}} {\color[HTML]{000000}\textbf{a}} {\color[HTML]{000000}\textbf{photo}} {\color[HTML]{000000}\textbf{for}} {\color[HTML]{000000}\textbf{inspiration}}{\color[HTML]{000000}\textbf{.}} {\color[HTML]{000000}\textbf{Can}} {\color[HTML]{000000}\textbf{you}} {\color[HTML]{000000}\textbf{show}} {\color[HTML]{000000}\textbf{me}} {\color[HTML]{000000}\textbf{two}} {\color[HTML]{000000}\textbf{pictures}} {\color[HTML]{000000}\textbf{of}} {\color[HTML]{000000}\textbf{flowers}} {\color[HTML]{000000}\textbf{and}} {\color[HTML]{000000}\textbf{describe}} {\color[HTML]{000000}\textbf{them}}{\color[HTML]{000000}\textbf{?}} 

\vspace{1em}
{\color[HTML]{000000}In} {\color[HTML]{000000}this} {\color[HTML]{000000}first} {\color[HTML]{000000}image}{\color[HTML]{bc7892},} {\color[HTML]{600734}a} {\color[HTML]{943b61}beautiful} {\color[HTML]{3e0f1f}arrangement} {\color[HTML]{550b2d}of} {\color[HTML]{030001}pink} {\color[HTML]{0a0205}flowers} {\color[HTML]{b47086}is} {\color[HTML]{19060c}on} {\color[HTML]{060103}display}{\color[HTML]{570a2f},} {\color[HTML]{510c2a}with} {\color[HTML]{e1b3ca}a} {\color[HTML]{530b2c}total} {\color[HTML]{550b2d}of} {\color[HTML]{5e0833}six} {\color[HTML]{030001}flowers} {\color[HTML]{bc7892}spread} {\color[HTML]{19060c}out} {\color[HTML]{0f0307}in} {\color[HTML]{570a2f}a} {\color[HTML]{bb7790}visually} {\color[HTML]{e1b3ca}appealing} {\color[HTML]{110408}manner}{\color[HTML]{030001}.}{\color[HTML]{210810}\textless{}\textbar{}image\textbar{}\textgreater{}}{\color[HTML]{6e023e}The} {\color[HTML]{c27e9b}roses} {\color[HTML]{000000}are} {\color[HTML]{6e023e}delicate} {\color[HTML]{6e023e}and} {\color[HTML]{210810}have} {\color[HTML]{b47086}a} {\color[HTML]{630636}soft}{\color[HTML]{1c070e},} {\color[HTML]{a2566f}pink} {\color[HTML]{b8748d}hue}{\color[HTML]{030001},} {\color[HTML]{c682a2}which} {\color[HTML]{550b2d}adds} {\color[HTML]{060103}to} {\color[HTML]{9f506c}the} {\color[HTML]{550b2d}overall} {\color[HTML]{e2b4cb}charm} {\color[HTML]{b36f85}of} {\color[HTML]{590930}the} {\color[HTML]{020001}scene}{\color[HTML]{260913}.} {\color[HTML]{0c0306}The} {\color[HTML]{19060c}bouquet} {\color[HTML]{020001}seems} {\color[HTML]{69043a}to} {\color[HTML]{000000}be} {\color[HTML]{18060c}the} {\color[HTML]{8c2a59}main} {\color[HTML]{c884a4}focus} {\color[HTML]{590930}of} {\color[HTML]{5e0833}the} {\color[HTML]{020001}image}{\color[HTML]{8a2657},} {\color[HTML]{b9758e}with} {\color[HTML]{19060c}the} {\color[HTML]{020001}flowers} {\color[HTML]{3e0f1f}captured} {\color[HTML]{550b2d}in} {\color[HTML]{5e0833}a} {\color[HTML]{020001}close}{\color[HTML]{570a2f}-up} {\color[HTML]{370d1b}shot}{\color[HTML]{110408},} {\color[HTML]{030001}showcasing} {\color[HTML]{4e0d28}their} {\color[HTML]{640637}intricate} {\color[HTML]{d193b2}details}{\color[HTML]{550b2d}.} {\color[HTML]{441022}The} {\color[HTML]{8a2757}display} {\color[HTML]{e3b7cd}creates} {\color[HTML]{cc89ab}an} {\color[HTML]{000000}elegant} {\color[HTML]{18060c}and} {\color[HTML]{530b2c}serene} {\color[HTML]{7a0647}atmosphere}{\color[HTML]{9e4f6b},} {\color[HTML]{490f25}ev}{\color[HTML]{a2566f}oking} {\color[HTML]{070103}a} {\color[HTML]{2c0b16}sense} {\color[HTML]{550b2d}of} {\color[HTML]{0c0306}romance} {\color[HTML]{310c18}and} {\color[HTML]{030001}appreciation} {\color[HTML]{a1546e}for} {\color[HTML]{19060c}the} {\color[HTML]{030001}beauty} {\color[HTML]{a2566f}of} {\color[HTML]{90325d}nature}{\color[HTML]{4e0d28}.} {\color[HTML]{020001}This} {\color[HTML]{620636}image} {\color[HTML]{19060c}features} {\color[HTML]{EECCDD}a} {\color[HTML]{000000}close}{\color[HTML]{000000}-up} {\color[HTML]{3e0f1f}view} {\color[HTML]{1c070e}of} {\color[HTML]{030001}a} {\color[HTML]{4e0d28}beautiful} {\color[HTML]{640637}pink} {\color[HTML]{a45a71}flower} {\color[HTML]{19060c}sitting} {\color[HTML]{030001}in} {\color[HTML]{0a0205}a} {\color[HTML]{060103}glass} {\color[HTML]{4a0e26}vase} {\color[HTML]{872154}on} {\color[HTML]{110408}a} {\color[HTML]{030001}table}{\color[HTML]{000000}.}{\color[HTML]{560a2e}\textless{}\textbar{}image\textbar{}\textgreater{}}{\color[HTML]{e1b3ca}The} {\color[HTML]{030001}flower} {\color[HTML]{000000}appears} {\color[HTML]{000000}to} {\color[HTML]{6e023e}be} {\color[HTML]{6e023e}a} {\color[HTML]{210810}ger}{\color[HTML]{4a0e26}ber}{\color[HTML]{8d2c5a}a}{\color[HTML]{590930},} {\color[HTML]{5e0833}with} {\color[HTML]{030001}a} {\color[HTML]{000000}vibrant} {\color[HTML]{5e0833}pink} {\color[HTML]{0f0307}color} {\color[HTML]{550b2d}and} {\color[HTML]{5e0833}a} {\color[HTML]{030001}prominent} {\color[HTML]{000000}center}{\color[HTML]{060103},} {\color[HTML]{580a2f}possibly} {\color[HTML]{a2566f}a} {\color[HTML]{030001}bud}{\color[HTML]{000000}.} {\color[HTML]{000000}The} {\color[HTML]{550b2d}glass} {\color[HTML]{000000}vase} {\color[HTML]{030001}is} {\color[HTML]{000000}positioned} {\color[HTML]{000000}at} {\color[HTML]{18060c}the} {\color[HTML]{000000}base} {\color[HTML]{560a2e}of} {\color[HTML]{510c2a}the} {\color[HTML]{6d033d}flower}{\color[HTML]{2c0b16},} {\color[HTML]{020001}enhancing} {\color[HTML]{000000}the} {\color[HTML]{19060c}visual} {\color[HTML]{020001}appeal} {\color[HTML]{4a0e26}of} {\color[HTML]{550b2d}the} {\color[HTML]{000000}scene}{\color[HTML]{020001}.} {\color[HTML]{1b060d}The} {\color[HTML]{380e1c}table} {\color[HTML]{19060c}beneath} {\color[HTML]{020001}the} {\color[HTML]{69043a}vase} {\color[HTML]{000000}provides} {\color[HTML]{18060c}a} {\color[HTML]{000000}simple} {\color[HTML]{790446}yet} {\color[HTML]{020001}elegant} {\color[HTML]{560a2e}backdrop}{\color[HTML]{15050a},} {\color[HTML]{030001}allowing} {\color[HTML]{1e070f}the} {\color[HTML]{1d070e}flower} {\color[HTML]{490f25}to} {\color[HTML]{000000}be} {\color[HTML]{550b2d}the} {\color[HTML]{510c2b}focal} {\color[HTML]{020001}point} {\color[HTML]{4a0e26}of} {\color[HTML]{590930}the} {\color[HTML]{5e0833}image}{\color[HTML]{020001}.}
& 
\includegraphics[width=1.0\linewidth]{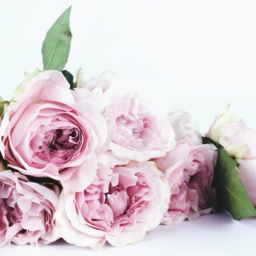} %
\includegraphics[width=1.0\linewidth]{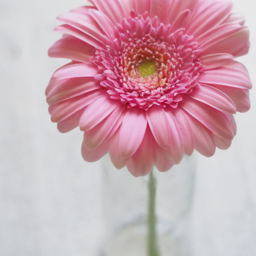}
\\

\bottomrule
\end{tabular}
}
\caption{\textbf{Concurrent interleaved text \& image generation.} OneFlow can insert variable number of images in the generated sequence, which are concurrently denoised alongside the text. This allows the text and images to depend on each other during the generation process.}
\label{fig:color_coded_example}
\end{figure} 

\section{Introduction}

Native Multimodal Models — models capable of handling both multimodal understanding and generation within a single backbone — have advanced considerably in visual understanding and generation. These models typically employ a unified transformer architecture with next-token prediction to handle both discrete and continuous generation~\citep{team2024chameleon, janus, janusflow, bagel, transfusion}. Recent work like Transfusion~\citep{transfusion} and Bagel~\citep{bagel} demonstrates that leveraging modality-specific training objectives within shared architectures can significantly improve performance, particularly on continuous modalities such as vision.

However, both autoregressive (\AR) and diffusion-based multimodal approaches face fundamental architectural constraints. Autoregressive models can handle interleaved data but require strict sequential generation — each image must be fully completed before text generation can continue, preventing simultaneous cross-modal refinement. Conversely, diffusion-based multimodal models such as MMaDA~\citep{mmada}, FUDOKI~\citep{fudoki}, and Unidisc~\citep{unidisc} enable simultaneous mixed-modal generation but only for predetermined single text-image pairs where modality assignments must be known a priori and rely on independent time schedules for each modality. Neither paradigm supports the simultaneous generation of variable-length interleaved sequences.

We present \textbf{\EF}, the first model to achieve simultaneous generation of interleaved data. Unlike autoregressive models that enforce sequential completion of each modality, and unlike diffusion models restricted to fixed length generation, \EF combines an insertion-based discrete text generation using Edit Flows with Flow Matching for image generation. This enables concurrent refinement of both text and images with per-image time schedules, using a novel interleaved time schedule.

Through controlled experiments across various model sizes and compute regimes, we demonstrate that \EF outperforms both autoregressive (\AR) and diffusion baselines on generation and understanding tasks while requiring 50\% fewer training FLOPS. Moreover, we find that concurrent mixed modal pretraining yields 4\% relative improvement on VQA and 1.5\% on image generation over sequential pretraining. We summarize our contributions below.


\begin{tcolorbox}[
    colback=blue!5,     
    colframe=teal!60!black,     
    arc=5pt,                    
    boxsep=5pt,                 
    left=5pt,                   
    right=10pt,                 
    top=5pt,                    
    bottom=5pt,                 
    boxrule=0.8pt,              
    drop shadow=gray!50!white,  
    enhanced jigsaw             
]
\textbf{Contributions:}
\begin{enumerate}[leftmargin=15pt]
    \item We introduce \EF, a non-autoregressive multimodal model that unifies image and text generation under a simultanous Edit Flow and Flow Matching framework.
    \item \EF enables new capabilities such as concurrent mixed-modal generation, which helps boost performance over uni-modal generation on a wide range of benchmarks.
    \item Through controlled experiences from 1B to 8B, we find that \EF scales better than autoregressive multimodal models, more so with mixed-modal training.
    \item \EF outperforms or is competitive with existing AR and diffusion-based models, on a diverse range of image generation and image understanding benchmarks.
\end{enumerate}
\end{tcolorbox}


\section{OneFlow: Mixed-Modal Generation through Flow Matching}

OneFlow handles multiple modalities through a sequence model, where elements in the sequence can be discrete tokens or continuous embeddings, \eg, of images. Concretely, let $\gT$ denote the space of a single element of the sequence, which can take either a discrete value, up to some fixed vocabulary size $M$, or a continuous value, \ie, $\gT = [M] \cup \R$.
Then our state space is defined as the set of all possible sequences up to some maximum length $N$, \ie, $\gX = \bigcup_{n=0}^N \gT^n$. 

During generation, our model transforms noisy sequences into clean sequences. We do this by combining discrete-valued and continuous-valued generative processes. Specifically, we make use of the Edit Flows~\citep{havasi2025edit} framework which enables variable-length sequence generation through the use of edit operations. It starts with a noisy sequence and iteratively applies edits until it is denoised into a generation. We focus on the \emph{insertion} capabilities of Edit Flows, which is conceptually simple yet extremely powerful, as it allows inserting arbitrary number of tokens---and images---into the generated sequence. When images are inserted, we initialize them with noise and then use Flow Matching~\citep{lipman2024flow} to generate the image. Since the same model predicts both the text edits and the image denoising, OneFlow achieves variable-length, non-autoregressive joint image and text generation. In the following, we state equations with only intuitive justifications and explanations. Full mathematical details and derivations can be found in Appendix~\ref{app:edit_flow_derivation}.

\subsection{Discrete Text Generation via Edit Flows}

Edit Flows uses a continuous-time Markov chain (CTMC) to iteratively refine variable-length discrete sequences. We start with an empty sequence $X_0 = \emptyset$ at time $t=0$, and transform the sequence through insertion operations. 
Let $\ins(x, i, a)$, $x \in \gX, i \in \{1, \dots, n\}, a \in \gT$, be the sequence resulting from inserting the token value $a$ to the right of position $i$ of the sequence $x$, resulting in
\begin{equation}
    \ins(x, i, a) = (x^1,\dots, x^i, a, x^{i+1}, \dots, x^{n}).
\end{equation}
This forms the primitive operation that we use during generation. 

During training, we take a data sequence $X_1$ and randomly delete tokens with equal probability to obtain $X_t$. This defines the process $X_{[0,1]}$ that we will fit to. The probability of each token being deleted is set by a monotonic scheduler $\kappa_t$ with $\kappa_0=0, \kappa_1=1$.
\begin{equation}\label{eq:schedule}
    \P(x^i \text{ in } X_t) = \kappa_t, \qquad \text{ for each }x^i \in X_1.
\end{equation}
In preliminary experiments, we tested different $\kappa_t$ but found that the linear schedule $\kappa_t = t$ works most consistently across our diverse benchmarks. \textit{Deleted tokens are removed from the sequence}. Noting that on average we retain $\E_t[\kappa_t]$ fraction of the original tokens, with the linear schedule we retain $50\%$ of the data sequence. This can lead to significant FLOPS savings during training, and tuning the scheduler can save even more if desired. 

\subsubsection{Parameterization} The parameterization of an Edit Flow model for insertions naturally decomposes into two predictions: (i) how many tokens are missing at the right of position $i$, and (ii) which tokens are missing. 
Thus, at each position $i$ of the sequence, our model outputs two quantities 
\begin{itemize}
    \item $\lambda^i: \mathcal{X} \rightarrow \R^+$ is a scalar that predicts the \emph{number of missing tokens} between $i$ and $i+1$.
    \item $Q^i : \mathcal{X} \rightarrow [M]$ is a normalized distribution that predicts \emph{what tokens are missing}.
\end{itemize}
These two predictions form the CTMC rate, so for sampling we would use (up to $o(h)$ error),
\begin{equation}\label{eq:sampling_step}
    \P\left(X_{t+h} = \ins(X_t, i, a) \;|\; X_t \right) = h \tfrac{\dot{\kappa}_t}{1 - \kappa_t} \lambda^i(X_t) Q^i (a | X_t).
\end{equation}
The ratio $\tfrac{\dot{\kappa}_t}{1 - \kappa_t}$ dictates the distribution of insertion times according to the schedule $\kappa_t$ imposed during training \eqref{eq:schedule}, where $\dot{\kappa}_t = \tfrac{d\kappa_t}{dt}$.
Note that unlike prior work \citep{havasi2025edit}, we factor out this ratio $\tfrac{\dot{\kappa}_t}{1 - \kappa_t}$ from the rate predictions and use a simplified model that is independent of $t$. In our practical implementation, we do not feed time values into the network for predicting insertions. 
While not theoretically justified, we found this $t$-independence assumption to work better in practice, likely because $X_t$ already contains sufficient information for predicting the insertions.

\subsubsection{Insertion prediction $\lambda^i$} The main component that determines whether insertions occur is the prediction head $\lambda^i$, which is trained by regressing onto the number of missing tokens. 
Each position $i$ of the noisy sequence $X_t$ has a corresponding number of missing tokens $k^i$, which is the number of deleted tokens between $X_t^i$ and $X_t^{i+1}$. 
The original Edit Flows loss was constructed through a choice of Bregman divergence \citep{holderrieth2024generator} which results in
\begin{equation}\label{eq:poisson_nll}
    \ell_\text{Poisson}(\lambda^i) = \tsum_i \lambda^i(X_t) - k^i \log \lambda^i(X_t).
\end{equation}
Alternatively, \eqref{eq:poisson_nll} can be interpreted as the negative log-likelihood of a Poisson distribution, so $\lambda^i$ is trained to fit a Poisson distribution to model missing token counts $k^i$. However, the distribution of $k$ has a very high concentration for $k^i=0$. Furthermore, during sampling, the key prediction is whether the missing token count is either zero or nonzero.
As such, we explicitly model the probability of inserting zero tokens.
\begin{equation}
\P(k = 0) = \pi, \quad \text{ and } \quad \P(k) = (1 - \pi) \text{Pois}(k ; \lambda_{\text{nonzero}} \;|\; k > 0) \quad \text{ for $k > 0$}
\end{equation}
\noindent
where $\pi \in (0, 1)$ is the probability of observing zero, and $\lambda_{\text{nonzero}} > 0$ is the rate parameter but restricted to only modeling the distribution of nonzero counts. 
We train $\pi$ by using a binary cross entropy (BCE) loss to detect if the missing count is zero, and we train $\lambda_{\text{nonzero}}$ using the original loss (\eqref{eq:poisson_nll}) on nonzero counts.
For sampling \eqref{eq:sampling_step}, we can use the expectation $\lambda^i(X_t) = (1-\pi^i(X_t)) \lambda^i_{\text{nonzero}}(X_t)$. 
However, we found that a consistently better sampling strategy is to first sample whether there are zero insertions using $\pi$, then simply use the rate $\lambda^i_{\text{nonzero}}(X_t)$ if there are nonzero insertions. We provide sampling pseudocode in Algorithm~\ref{alg:sampling}.\looseness=-1

\subsubsection{Bag-of-tokens prediction $Q^i$} To determine what token to insert at each position, we make use of the output head $Q$ which is a softmax over the discrete vocabulary $[M]$. We use the same Edit Flows loss, which is a sum of cross-entropy loss. Let $\mathcal{A}_i$ denote the set of deleted tokens between $X_t^i$ and $X_t^{i+1}$, then for each position $i$, the loss is
\begin{equation}\label{eq:token_loss}
    \ell_{\text{tokens}}(Q^i) = - \tsum_{a \in \mathcal{A}_i} \log Q^i(a | X_t).
\end{equation}

\subsubsection{Combined loss} At each training iteration, we randomly delete tokens from the data sequence, and learn to predict the set of missing tokens at each position, resulting in the total insertion loss:
\begin{equation}\label{eq:insertion_loss}
    \mathcal{L}_{\text{text}} = \E_{t, X_t | X_1} \left[ \frac{1}{n} \sum_{i=1}^{n} \ell_{\text{tokens}}(Q^i) + \ell_\text{Poisson}(\lambda^i_\text{nonzero}) \1_{[k_i > 0]} + \ell_\text{BCE}(\pi^i) \right]
\end{equation}
where $n$ is the length of the noisy sequence $X_t$. Note that this differs from the original training objective in Edit Flows \citep{havasi2025edit} which additionally weights the loss by the factor $\tfrac{\dot{\kappa}_t}{1 - \kappa_t}$, not affecting the optimal solution. We found that not using this factor produces better results.

\subsection{Continuous Image Generation via Flow Matching}

Following standard practice, we generate images starting from a Gaussian noise of fixed dimension $N_{\text{img}}$, applying a deterministic generation procedure that follows an ordinary differential equation. Let $Y_t \in \R^{N_{\text{img}}}$ denote the noisy image, then the generative process is
\begin{equation}
    \tfrac{d}{dt} Y_t = v(Y_t, t), \qquad Y_0 \sim \mathcal{N}(0, I),
\end{equation}
where $v: \R^{N_{\text{img}}}\times\R \rightarrow \R^{N_{\text{img}}}$ is a velocity field that determines the direction to transform $Y_t$ into a clean sample by $t=1$.
During training, we sample a noise $Y_0$ and obtain $Y_t$ with a linear schedule $Y_t = tY_1 + (1-t)Y_0$. The Flow Matching loss can then be written as
\begin{equation}\label{eq:flow_matching}
    \mathcal{L}_\text{image} = \E_{t, Y_0, Y_1} \left\lVert v(Y_t, t) - (Y_1 - Y_0) \right\rVert^2.
\end{equation}

In OneFlow, we use a pretrained autoencoder to map images into latent space. We then design the velocity network $v(\cdot)$ to use a shared Transformer backbone as text but with additional U-Nets to downsample and upsample between the backbone and autoencoder embedding spaces, making use of the same architectural design as Transfusion~\citep{transfusion}. See illustration in Figure \ref{fig:architecture}.

\subsection{Concurrent Mixed-modal Generation}\label{sec:mixed_modal_schedules}

To generate multiple modalities, we simply concatenate them into a single sequence. We now present two multimodal time schedules, an independent schedule that can be used when the number of images is known, and an interleaved schedule that needs to be used when the number of images is arbitrary. OneFlow is designed to work with variable-length text and variable number of images.

\subsubsection{Independent mixed-modal generation}
We can consider the simple case with a fixed number of images---typically one. In such case, we can generate both the text and image simultaneously by using two time values $t_\text{text}$ and $t_\text{img}$, where $t_\text{text}$ determines the state of the insertion generation process and $t_\text{img}$ determines the image generation process. 
Following prior work, we simply set independent time schedules, one for the text and one for each image. This allows the modalities to be concurrently generated and be dependent on each other during the generation process. However, this na\"ive process does not allow us to insert images.

\subsubsection{Interleaved mixed-modal generation}
A much more complicated setting arises when the number of images is variable and images are being inserted as part of the generation process. Similar to the text-only setting, we start generating from the empty sequence.
We then model image insertion as a special token value \image, which is added to the token prediction output $Q$.
During generation, when the model predicts an image insertion, we insert noise embeddings of dimension $N_{\text{img}}$ into the sequence to represent an inserted image initialized at $t_\text{img} = 0$. 
\begin{equation}
    \ins(x, i, \image) = \big(x^1,\dots, x^i, y^1, \dots, y^{N_\text{img}}, x^{i+1}, \dots, x^{n}\big), \qquad y^i \sim \mathcal{N}(0, I).
\end{equation}
Subsequent steps during generation would then simultaneously generate the image embeddings while also performing more insertions into the sequence. However, since the image is generated at a later time, this implies there is a delay between the image time and the text time, \ie $t_\text{img} \leq t_\text{text}$, which needs to be taken into account during training.

During training, we need to ensure that the text and image noise levels are consistent with with the ones seen during generation. 
Based on the schedule in \eqref{eq:schedule}, the time at which an insertion happens is a random variable that has $\kappa$ as its cumulative density function, so the time difference between the inserted image time $t_\text{img}$ and the initial text time $t_\text{text}$ is given by
\begin{equation}\label{eq:interleaved_time_schedule}
    t_\text{img} = t_\text{text} - \kappa^{-1}(u), \qquad \text{ where } u \sim \text{Unif(0, 1)}.
\end{equation}
We call this the \emph{interleaved time schedule}, which imposes a distributional dependency between the time values $t_\text{img}$ for each image and the text time $t_\text{text}$. 
In order for the model to learn to fully generate all images, during training we sample from an extended time interval, $\tau_{\text{text}}$ from $[0, 2]$, since the \image token can be inserted at $\tau_{\text{text}}=1.0$ at the latest, and fully denoised by $\tau_{\text{text}}=2.0$. The probability for each token being in $X_t$ is then determined by $\kappa(\min\{1, \tau_{\text{text}}\})$ in place of \eqref{eq:schedule}.
We also sample for each image an extended time value $\tau_{\text{img}} = \tau_\text{text} - \kappa^{-1}(u)$. 
Finally, we determine if an image is deleted from the sequence by checking $\tau_{\text{img}} < 0$, and if so, the insertion loss (\eqref{eq:insertion_loss}) will include the \image{} token which the model would learn to insert. Otherwise if $\tau_{\text{img}} \geq 0$, the image is in the sequence and we set $t_\text{img} = \min\{1, \tau_{\text{img}}\}$; using the Flow Matching loss \eqref{eq:flow_matching} to train the velocity. A detailed derivation and in-depth explanation can be found in Appendix \ref{app:interleaved_schedule}.

\section{Findings and Unlocking New Capabilities in Unified Multimodal}

To study the performance of \EF{} against \AR, we design a series of controlled experiments. We establish strong baselines by comparing both our approach and \AR{} against other unified multimodal models in the literature. Finally, we explore the novel capabilities that \EF{} enables beyond existing methods. We present our experimental results through five research questions:\par
\begin{tabular}{@{}ll}
    \textcolor{blue}{\S\ref{controlled}}      & How does \EF{} scale compared to \AR? \\
    \textcolor{blue}{\S\ref{sec:mixed_modal}} & What is the impact of mixed-modal vs. sequential pretraining? \\
    \textcolor{blue}{\S\ref{sec:vqa_sampling}}& What emergent behaviors does \EF{} exhibit during generation?\\
    \textcolor{blue}{\S\ref{sec:comparison} }  & How does \EF{} compare to other unified multimodal models? \\
    \textcolor{blue}{\S\ref{sec:new_capabilities}} & What new capabilities does \EF{} enable beyond existing methods? \\
\end{tabular}

\subsubsection{Training stages}
Our training consists of two main stages: multimodal pretraining and instruction finetuning. During the pretraining stage, we use a mixture of image understanding and image generation data to learn representations for both image and text. We trained with a sequence length of 512 and a global batch size of 4096. We set the mixed generation probability (the likelihood of concurrently generating clean text and the image) to be either 0 or 0.2.

For finetuning, we use a mixture of VQA, text, and interleaved data to give the model the ability to respond to visual question answering problems. We also fine-tune on image generation data at a higher resolution of 512$\times$512 to improve the model’s image generation capabilities. We study the model's behavior at the 1B scale for our ablations and controlled experiments, and the scaling trend up to 8B is detailed in Section~\ref{controlled}.

\subsubsection{Datasets}
For multimodal pretraining, we use image-text pairs from a filtered version of the Conceptual Captions dataset (CC12M \citep{sharma2018conceptual}), the YFCC dataset \citep{thomee2016yfcc100m}, and licensed data, for a total of 400M examples. 
During instruction finetuning, we use a filtered image portion of the PerceptionLM dataset \citep{cho2025perceptionlm}, interleaving data from Chameleon \citep{team2024chameleon}, and a filtered portion of the Cambrian-7M~\citep{tong2024cambrian} dataset, totaling around 40M examples.

\subsubsection{Baselines} To evaluate our model's performance against existing methods, we compare against two baselines: (1) an autoregressive (\AR) + Flow Matching (FM) multimodal model based on Transfusion~\citep{transfusion}, where text tokens are generated autoregressively and image tokens via FM, and (2) a masked diffusion model based on LLaDA~\citep{nie2025large}. For the masked diffusion baseline, we tested two sampling variants: low-confidence and random remasking, with random remasking performing better across all experiments. Unlike Transfusion, we follow Janus-Flow~\citep{janusflow} and adopt a dual-encoder setup. For image encoders, we use a pretrained SigLIP2 ViT-SO400M-16@512~\citep{siglip2} for understanding and an SD3 VAE~\citep{sd3} for generation. Following Transfusion, we use U-Net adapters. Figure \ref{fig:architecture} illustrates how \EF consumes a mixed-modal sequence. To improve performance of the AR model, we follow Chameleon~\citep{team2024chameleon} and sweep over several generation lengths for optimizing CIDEr score.
For image generation, we use a first order Euler solver with entropy rectifying guidance~\citep{ifriqi2025entropyrectifyingguidancediffusion}, we set the guidance scale to $5.0$ across $50$ sampling steps.

\subsubsection{Model Configurations}
To investigate scaling trends, we train models at three different sizes – 1B,
3B, and 8B parameters, and token counts. For each benchmark, we plot all results over log-FLOPS curve and regress linear trendlines. Following Transfusion, we also estimate relative compute efficiency by measuring the parity FLOPS ratio: the ratio between the number of FLOPS required by our \AR and \EF to reach the same level of performance. 

\subsubsection{FLOPS Estimation}
To estimate the FLOPS during training, we track the number of tokens throughout training for all models. During multimodal pretraining, we do not delete images and \EF uses on average 50\% fewer generated \textit{text} tokens compared to \AR. 
We use Flash Attention and follow their FLOPS estimation \citep{flashattention}.

\subsubsection{Evaluation setup}
Following Cambrian~\citep{tong2024cambrian} and PLM~\citep{cho2025perceptionlm}, we group VQA tasks into five groups: General, Knowledge, OCR \& Chart, Hard Perception, and Hallucination. We evaluate image generation quality using the FID metric~\citep{fid} on the COCO-2014~\citep{coco2014} validation set, using our base model at 256$\times$256 resolution. To assess prompt alignment, we report CLIPScore~\citep{clipscore} and DPG-Bench~\citep{dpgbench}. Additionally, we include WISE~\citep{wise} cultural to better understand knowledge-based generation.

\subsection{\EF Scales Better than \AR}
\label{controlled}
\begin{figure*}
    \centering
    \includegraphics[width=\linewidth]{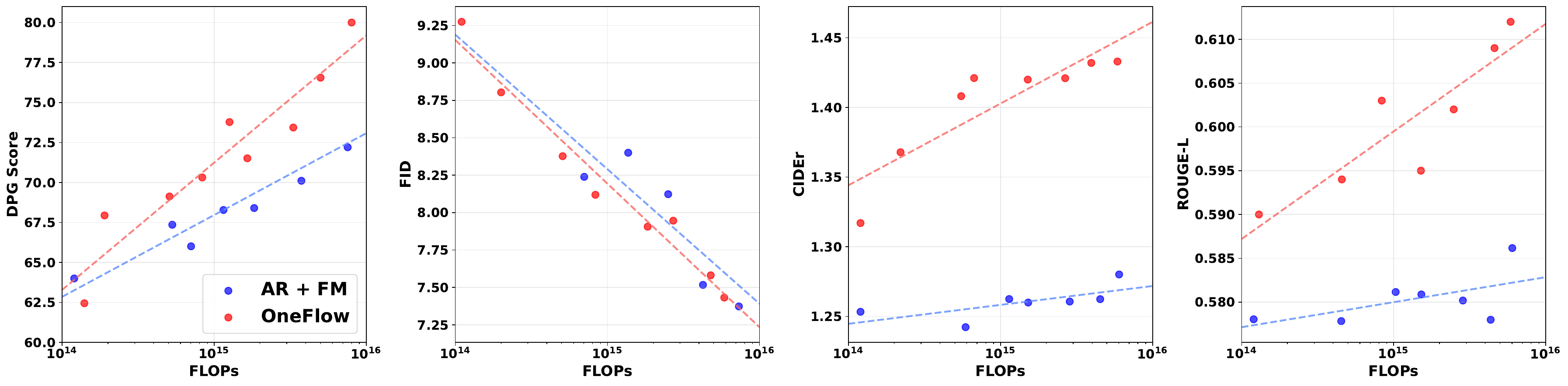}
    \caption{\textbf{Performance of \EF vs. \AR+FM baseline models at different model scales, data and compute.} For text-to-image generation, we report DPG-Bench and FID. For image-to-text caption quality, we report CIDEr and ROUGE.  In every benchmark, \EF consistently exhibits better scaling laws than \AR + FM. Model sizes include 1B, 3B, and 8B.}
    \label{fig:scaling_pretraining}
\end{figure*}

We first discuss controlled experiments, where we study the performance across different frameworks when trained on the same data sets using the same backbone architecture.
Specifically, we study the performance of \EF and \AR in controlled settings at various model sizes and token counts. To ensure \EF has no advantage in data-constrained settings, we trained both models on 2B image-text pairs over 500k iterations using a batch size of 4096. Both models were initialized from Llama 3.2 1B~\citep{llama3modelcard}. For AR, the number of tokens predicted during training equals the sequence length, whereas for OneFlow, the number of predicted tokens corresponds to the number of deleted tokens, which on average is 50\% of the data sequence. We measure compute efficiency using the parity FLOP ratio (Table \ref{tab:parity_ratio}), which quantifies the relative FLOPS required by \EF to match \AR's performance. Our results show that \EF achieves the largest compute savings on dense prompt alignment and captioning tasks, requiring less than half the FLOPS of \AR to reach the same level. 

We find that \EF scales better than \AR on every benchmark. This scaling advantage is especially pronounced on DPG Bench, where \EF scales significantly better than \AR. For image captioning as well, \EF shows a notable performance gap relative to \AR. Figure \ref{fig:scaling_pretraining} visualizes the scaling trend, and the final metrics are shown in Tables \ref{tab:pretraining_metrics} and \ref{tab:performance_comparison}, along with a comparison against other state-of-the-art models. A detailed breakdown of the controlled comparisons among VQA task categories can be found in Table \ref{tab:vqa_controlled_barplot}.

\begin{SCtable}[1]
\centering
\resizebox{.5\linewidth}{!}{
\begin{tabular}{lcccc}
\toprule
 & \textbf{DPG $\uparrow$} & \textbf{FID $\downarrow$} & \textbf{CIDEr $\uparrow$} & \textbf{ROUGE $\uparrow$} \\
\midrule
Parity FLOPS Ratio & 0.49 & 0.97 & 0.32 & 0.42 \\
\bottomrule
\end{tabular}}
\caption{\textbf{Parity FLOPS ratio in a controlled experiment.} Both models were trained on 2B image-text pairs. Parity FLOPS Ratio represents the relative amount of \EF FLOPS needed to match the final \AR + FM performance.}
\label{tab:parity_ratio}
\end{SCtable}

\subsection{Mixed Modal Pre-training Enables Better Generation and Understanding}
\label{sec:mixed_modal}

\begin{figure}
    \centering
    \includegraphics[width=\linewidth]{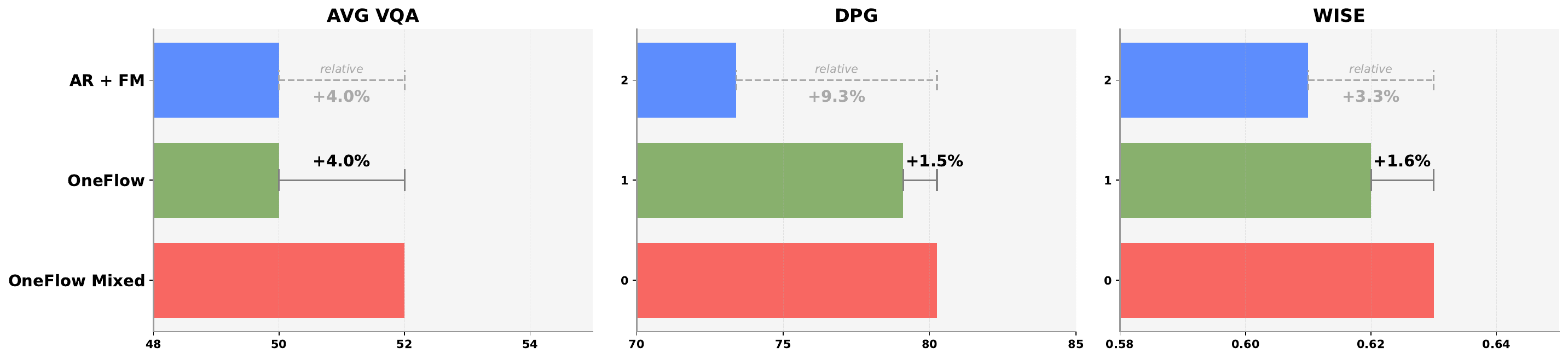}
    \caption{\textbf{Mixed modal vs Sequential pretraining.} Mixed modal pretraining vs sequential pretraining. Mixed pretraining achieves 4\% relative improvement on VQA tasks and slight improvements on image generation as well.}
    \label{fig:mixed_vs_clean}
    \vspace{-1em}
\end{figure}

\begin{wrapfigure}{r}{0.38\textwidth}
    \vspace{-20pt}
    \centering
    \includegraphics[width=0.36\textwidth]{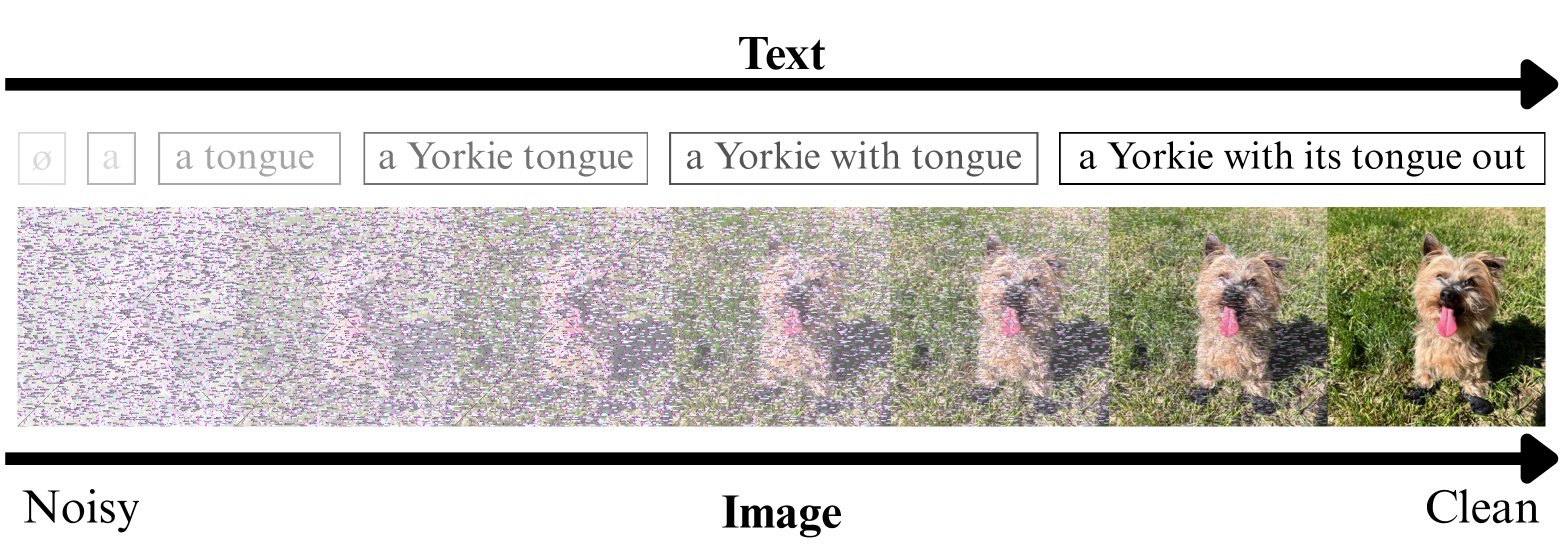}
    \caption{Mixed modal concurrent training.}
    \label{fig:mixed_gen}
    \vspace{-20pt}
\end{wrapfigure}

In this section, we study the impact of mixed-modal pretraining by comparing concurrent and sequential training approaches. Specifically, we train two 1B variants: one using sequential pretraining (either text-to-image or image-to-text), and another using the same data but with 20\% of examples incorporating concurrent generation.
Our ablation experiment reveals that mixed-modal training significantly improves downstream visual question answering performance (Figure~\ref{fig:mixed_vs_clean}) and provides a minor boost to image generation. These results demonstrate the benefits of concurrent mixed-modal pretraining. Given these consistent improvements, we use the mixed-modal pretrained model for all subsequent finetuning experiments.\looseness=-1

\subsubsection{Mixed Generation Probability}
\begin{figure}[H]
    \centering
    \includegraphics[width=\linewidth]{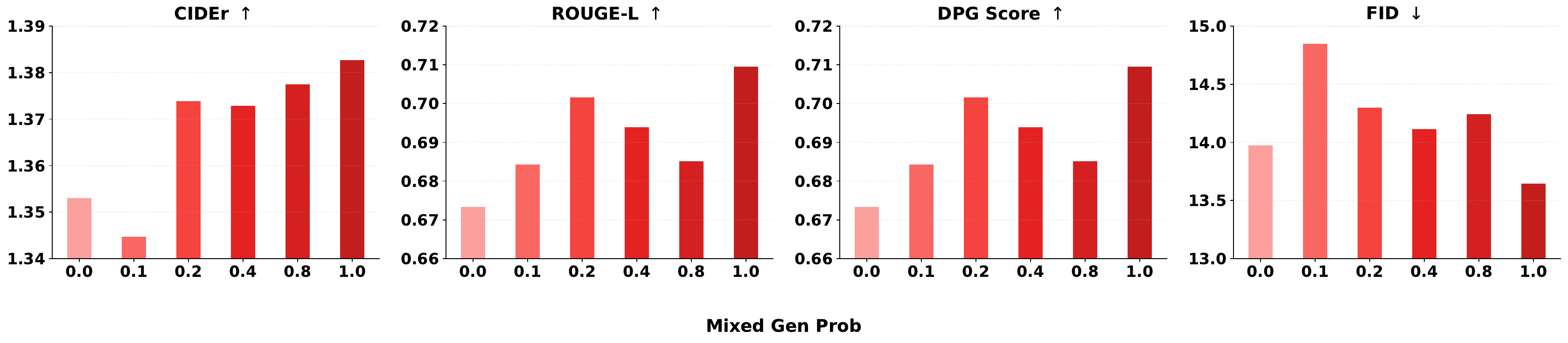}
    \caption{\textbf{Effect of mixed-generation probability on performance.} Higher mixed-generation probabilities consistently improve image understanding (CIDEr, ROUGE-L) and compositional generation (DPG Score), while image generation quality (FID) remains stable. This demonstrates that mixed training enhances understanding and compositional capabilities without compromising generation performance.}
    \vspace{-1em}
\label{fig:mixed_gen_ablation}
\end{figure}

In this section, we study the impact of different mixed-generation probabilities over 200k training steps. Results show a clear monotonic trend: higher mixed-generation probability consistently improves image understanding across all metrics. CIDEr scores increase from 135.3 (0\% mixed) to 138.3 (100\% mixed), ROUGE-L improves from 0.67 to 0.71, and DPG Score rises from 0.67 to 0.71. Importantly, image generation quality (FID, DPG Score) remains stable across all probabilities, demonstrating that mixed training strengthens understanding capabilities without degrading generation performance.\looseness=-1

\begin{table}
\renewcommand{\arraystretch}{1.05}
\centering
\scriptsize
\begin{adjustbox}{max width=\textwidth}
\begin{tabular}{@{}l c cc cccc ccc c@{}}
\toprule
& & & & \multicolumn{4}{c}{\cellcolor{colGen}\textbf{Image Generation}} & \multicolumn{3}{c}{\cellcolor{colHalluc}\textbf{Captioning}} \\
\cmidrule(lr){5-8} \cmidrule(lr){9-11} 
\textbf{Model} & \textbf{Size} & \rotatebox{45}{\textbf{Text}} & \rotatebox{45}{\textbf{Image}} & 
\cellcolor{colGen}\textbf{FID$\downarrow$} & 
\cellcolor{colGen}\textbf{CLIP$\uparrow$} & 
\cellcolor{colGen}\textbf{DPG$\uparrow$} & 
\cellcolor{colGen}\textbf{WISE (c.) $\uparrow$} & 
\cellcolor{colHalluc}\textbf{CIDEr$\uparrow$} & 
\cellcolor{colHalluc}\textbf{ROUGE$\uparrow$} &
\cellcolor{colHalluc}\textbf{BLEU4$\uparrow$} \\
\midrule
\multicolumn{11}{l}{\textit{Unified MLM}} \\
MetaMorph~\citep{metamorph}& 7B & AR & AR & 11.8 & \cellhi 26.6 & -- & -- & -- & -- & -- \\
LMFusion~\citep{lmfusion}  & 7B & AR & Diff & 14.0 & 24.4 & -- & -- & 38.4 & -- & -- \\
Transfusion~\citep{transfusion} & 7B & AR & Diff & 16.0 & 26.5 & 77.8 & -- & 33.7 & -- & -- \\
Janus-Pro~\citep{januspro} & 1.5B & AR & AR & \phantom{$^\dagger$}15.2$^\dagger$ & \phantom{$^\dagger$}26.0$^\dagger$ & \phantom{$^\dagger$}82.0$^\dagger$ & 0.20 & -- & -- & -- \\
Janus-Flow~\citep{janusflow} & 1.5B & AR & FM & \phantom{$^\dagger$}12.4$^\dagger$  & \phantom{$^\dagger$}26.1$^\dagger$ & \phantom{$^\dagger$}80.1$^\dagger$  & 0.13 & -- & -- & -- \\
Bagel~\citep{bagel} & 7B & AR & FM & \phantom{$^\dagger$}27.7$^\dagger$ & \phantom{$^\dagger$}26.2$^\dagger$ &  \cellhii \phantom{$^\dagger$}84.7$^\dagger$ & 0.44 & -- & -- & -- \\
\midrule
\multicolumn{11}{l}{\textit{Multimodal Diffusion}} \\
UniDisc~\citep{unidisc} & 1.4B & Mask & Mask & 23.9 & -- & -- & -- & -- & -- & -- \\
D-DiT~\citep{ddit}  & 2B & Mask & Diff & -- & -- & -- & -- & 56.2 & -- & -- \\
Muddit~\citep{muddit} & 1B & Mask & Mask & -- & -- & -- & -- & 59.7 & -- & --\\
MMaDA~\citep{mmada}  & 8B & Mask & Mask & \phantom{$^\dagger$}33.2$^\dagger$ & \phantom{$^\dagger$}25.1$^\dagger$ & \phantom{$^\dagger$}74.2$^\dagger$& \cellhi 0.67 & -- & -- & -- \\
FUDOKI~\citep{fudoki} & 1.5B & DFM & DFM & -- & -- & \cellhi 83.6 & -- & -- & -- & -- \\
\midrule
\multicolumn{11}{l}{\textit{\textbf{Controlled Comparisons}}} \\
AR + FM Ablation & 1B & AR & FM & 12.2 & 26.5 & 73.4 & 0.61 & 123.9 & 57.2 & 0.39 \\
Mask + FM Ablation & 1B & Mask & FM & 11.3 & 26.5 & 75.5 & 0.64 & 128.4 & 58.6 & 0.39 \\
\textbf{OneFlow} & 1B & EF & FM & 12.1 & \cellhi 26.6 & 79.1 & 0.62 & 138.1 & 60.8 & 0.41 \\
\textbf{OneFlow Mixed} & 1B & EF & FM & \cellhi 9.7 & \cellhi 26.6 & 80.3 & 0.63 & 139.8 & 60.9 & \cellhi 0.42 \\
\cmidrule{1-11}
\textbf{OneFlow} & 8B & EF & FM & 10.7 & \cellhii 26.7 & 79.3 & 0.65 & \cellhi 141.1 & \cellhii 61.1 & \cellhi 0.42 \\
\textbf{OneFlow Mixed} & 8B & EF & FM & \cellhii 9.5 & \cellhi 26.6 & 80.4 & \cellhii 0.68 & \cellhii 142.1 & \cellhii 61.1 & \cellhii 0.43 \\
\bottomrule
\end{tabular}
\end{adjustbox}
\caption{\textbf{Image generation and captioning benchmarks after multimodal pretraining.} We find that mixed-modal training consistently improves performance. $^{\dag}$Evaluated using official open-source model weights. Highlighting denotes best results across all models.}
\label{tab:pretraining_metrics}
\vspace{-15pt}
\end{table}
\begin{table}
\centering
\renewcommand{\arraystretch}{1.1}
\begin{adjustbox}{max width=\textwidth}
\begin{tabular}{
    l c
    c c c c
    c c
    c c c
    c
    c
}
\toprule
& & 
\multicolumn{4}{c}{\cellcolor{colGen}\textbf{General}} &
\multicolumn{2}{c}{\cellcolor{colKnow}\textbf{Knowledge}} &
\multicolumn{3}{c}{\cellcolor{colOCR}\textbf{OCR \& Chart}} &
\cellcolor{colVision}\textbf{Vision} &
\cellcolor{colHalluc}\textbf{Halluc.} \\
\cmidrule(lr){3-6} \cmidrule(lr){7-8} \cmidrule(lr){9-11} \cmidrule(lr){12-12} \cmidrule(lr){13-13}
\textbf{Model} & \rotatebox{45}{\textbf{Params}} 
& \cellcolor{colGen}\rotatebox{90}{\textbf{MMB}} 
& \cellcolor{colGen}\rotatebox{90}{\textbf{VQAv2}} 
& \cellcolor{colGen}\rotatebox{90}{\textbf{GQA}} 
& \cellcolor{colGen}\rotatebox{90}{\textbf{MME}} 
& \cellcolor{colKnow}\rotatebox{90}{\textbf{MMMU}} 
& \cellcolor{colKnow}\rotatebox{90}{\textbf{AI2D}} 
& \cellcolor{colOCR}\rotatebox{90}{\textbf{DocVQA}} 
& \cellcolor{colOCR}\rotatebox{90}{\textbf{ChartQA}} 
& \cellcolor{colOCR}\rotatebox{90}{\textbf{TextVQA}} 
& \cellcolor{colVision}\rotatebox{90}{\textbf{RealWorld}} 
& \cellcolor{colHalluc}\rotatebox{90}{\textbf{POPE}} 
\\
\midrule
\multicolumn{13}{l}{\; \textit{Multimodal LM}} \\
Show-O~\citep{showo} & 1.3B & -- & -- & 61.0 & 1232.9 & 27.4 & -- & -- & -- & -- & -- & 84.5 \\
MetaMorph~\citep{metamorph} & 7B & 75.2 & -- & -- & -- & 41.8 & -- & -- & 37.1 & 60.5 & 58.3 & -- \\
Janus-Flow~\citep{janusflow}& 1.5B & 74.9 & 79.8 & 60.3 & 1333.1 & 29.3 & -- & 64.6 & 55.5 & -- & -- & 88.0 \\
Janus-Pro$^\dagger$~\citep{januspro} & 1.5B & 73.4 & 67.9 & 59.3 & 1443.0 & 33.4 & 62.8 & 21.2 & 35.8 & 53.9 & 53.5 & 84.8 \\
Janus-Pro$^\dagger$~\citep{januspro} & 7B & 76.9 & 74.1 & 62.0 & 1531.0 & 38.2 & 68.1 & 24.3 & -- & 57.2 & 56.4 & 85.2 \\
\midrule
\multicolumn{13}{l}{\; \textit{Mask Diffusion}} \\
Muddit~\citep{muddit} & 1B & -- & 67.7 & 57.1 & 1104.6 & -- & -- & -- & -- & -- & -- & -- \\
D-DiT~\citep{ddit} & 2B & -- & 60.1 & 59.2 & 1124.7 & -- & -- & -- & -- & -- & -- & 84.0 \\
MMaDA~\citep{mmada} & 8B & 68.5 & 76.7 & 61.3 & -- & 30.2 & -- & -- & -- & -- & -- & 86.1 \\
\multicolumn{13}{l}{\; \textit{Discrete Flow}} \\
FUDOKI~\citep{fudoki}  & 1.5B & 73.9 & -- & 57.6 & 1485.4 & 34.3 & -- & -- & -- & -- & -- & 86.1 \\
\midrule
\multicolumn{13}{l}{\; \textit{\textbf{Controlled Comparisons}}} \\
AR + FM Ablation & 1B & 64.7 & 66.0 & 55.0 & 1394.8 & 28.9 & \cellhjj 59.1 & 22.7 & \cellhjj 35.5 & 48.3 & 50.5 & 85.6 \\
Mask + FM Ablation & 1B & 66.0 & 64.4 & 55.6 & 1462.2 & 28.4 & 55.1 & 18.7 & 34.7 & 44.6 & 50.6 & 84.6 \\
\textbf{\EF} & 1B & \cellhjj 69.0 & \cellhjj 67.7 &  \cellhjj 57.8 & \cellhjj 1497.1 & \cellhjj 29.8 & 58.5 & \cellhjj 23.8 & 35.0 & \cellhjj 50.4 & \cellhjj 50.6 & \cellhjj 85.7 \\
\cmidrule{1-13}
\textbf{\EF} & 8B & 72.5 & 73.7 & 61.9 & 1542.5 & 33.1 & 63.4 & 37.1 & 42.1 & 58.6 & 54.0 & 86.3 \\
\bottomrule
\end{tabular}
\end{adjustbox}
\caption{\textbf{VQA performance comparison.} \EF outperforms \AR and Mask models across all benchmarks in controlled experiments using identical finetuning data. Highlighting shows best results in the 1B controlled comparisons. Our results are also competitive with existing autoregressive and discrete diffusion models. $^{\dag}$Evaluated using official open source weights.}
\label{tab:performance_comparison}
\end{table}

\subsection{Hierarchical Generation Enables Emergent Reasoning}
\label{sec:vqa_sampling}

In Figure~\ref{fig:vqa_viz}, we present \EF's sampling process when prompted with a visual question. In response to prompts such as \textit{"\{question\} Explain why."}, \EF generates a reasoning chain before arriving at the final answer, without any Chain-of-Thought (CoT)~\citep{cot} prompting or RL post-training. For example, when asked \textit{"Is there a snowboard in the image? Explain why."}, the model first implicitly performs visual search by examining the image and searching through likely locations for the snowboard. Similarly, for the math puzzle in (Figure~\ref{fig:vqa_viz} bottom), the model first identifies objects in the image that match the prompt description—the green sphere and the large shiny cylinder—before arriving at the final answer.

\begin{figure*}
    \centering
    \begin{subfigure}{0.5\textwidth}
        \centering
        \includegraphics[width=\textwidth, height=0.35\textheight, keepaspectratio]{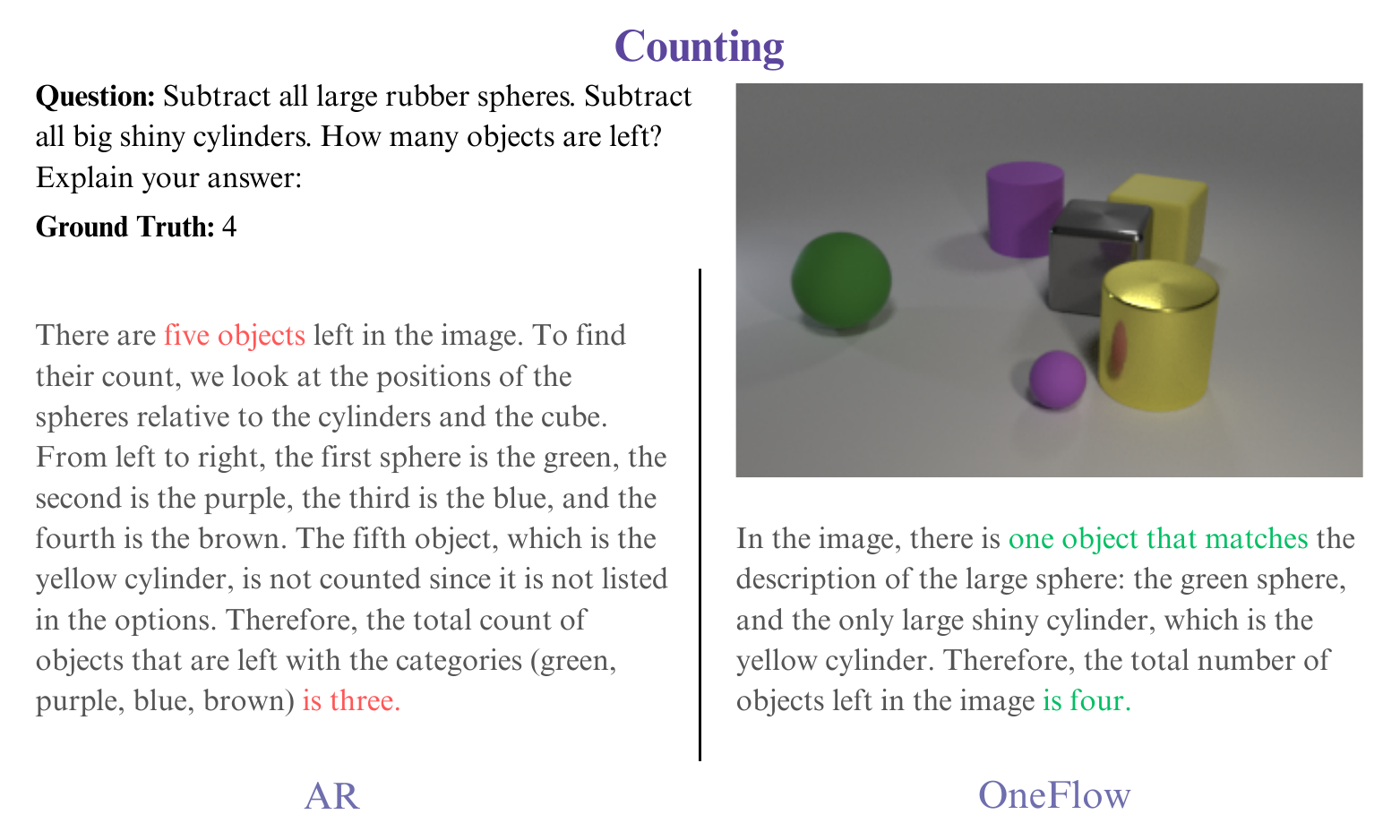}
        \label{fig:sub1}
    \end{subfigure}%
    \begin{subfigure}{0.5\textwidth}
        \centering
        \includegraphics[width=\textwidth, height=0.35\textheight, keepaspectratio]{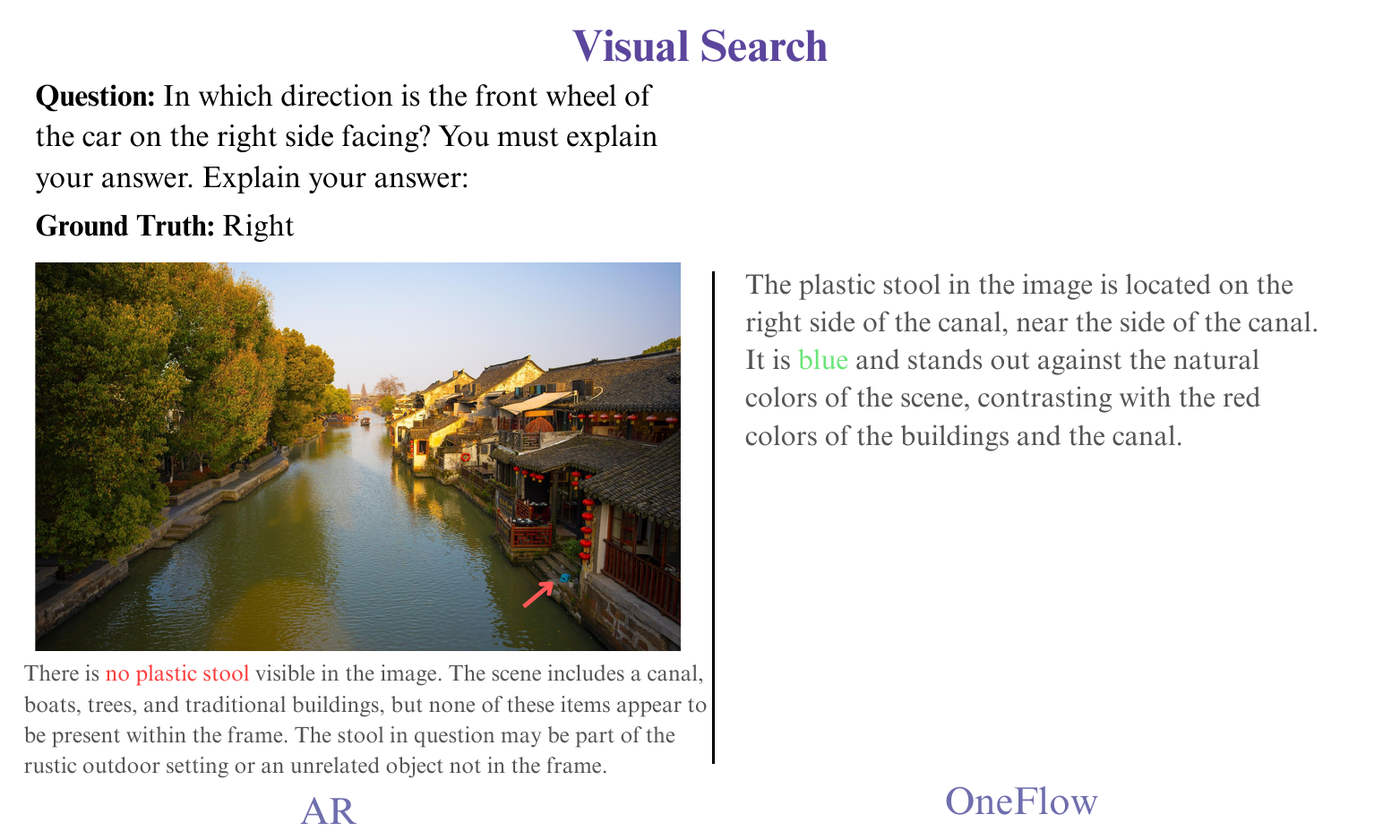}
        \label{fig:sub3}
    \end{subfigure}
    \vspace{-3em}
    \caption{
    (\textit{left}) \EF{} first locates the object in the prompt before calculation.
    (\textit{right}) \EF{} locates the target (stool), then analyzes color contrast. A red arrow is added for clarity and is not in the original image.}
    \label{fig:vqa_qualitative_comparison}
\end{figure*}

Our results align with findings in Physics of LLMs~\citep{physics_of_llm} and MetaMorph~\citep{metamorph}, where the authors suggest that LLMs precompute reasoning graphs before generating tokens. However, our findings demonstrate that the model can perform the same reasoning chain without autoregressive decoding. This suggests that reasoning capabilities can emerge in non-autoregressive architectures and transfer effectively to \EF. We show more example VQA generations compared to the AR baseline in Figure \ref{fig:vqa_qualitative_comparison}.

\begin{figure*}
    \tiny
    \centering

    \setlength\tabcolsep{1pt}      
    \renewcommand{\arraystretch}{1.0}
    \setlength\fboxsep{0pt}        
    \newcommand{\tightimg}[1]{\fbox{\includegraphics[width=22.1mm,height=22.1mm]{#1}}}

    \begin{adjustbox}{width=\textwidth}
    \begin{tabular}{c*{4}{p{28mm}}}
        \begin{sideways}\small{\;\;\;\;\;\;\; Bagel}\end{sideways} &
        \tightimg{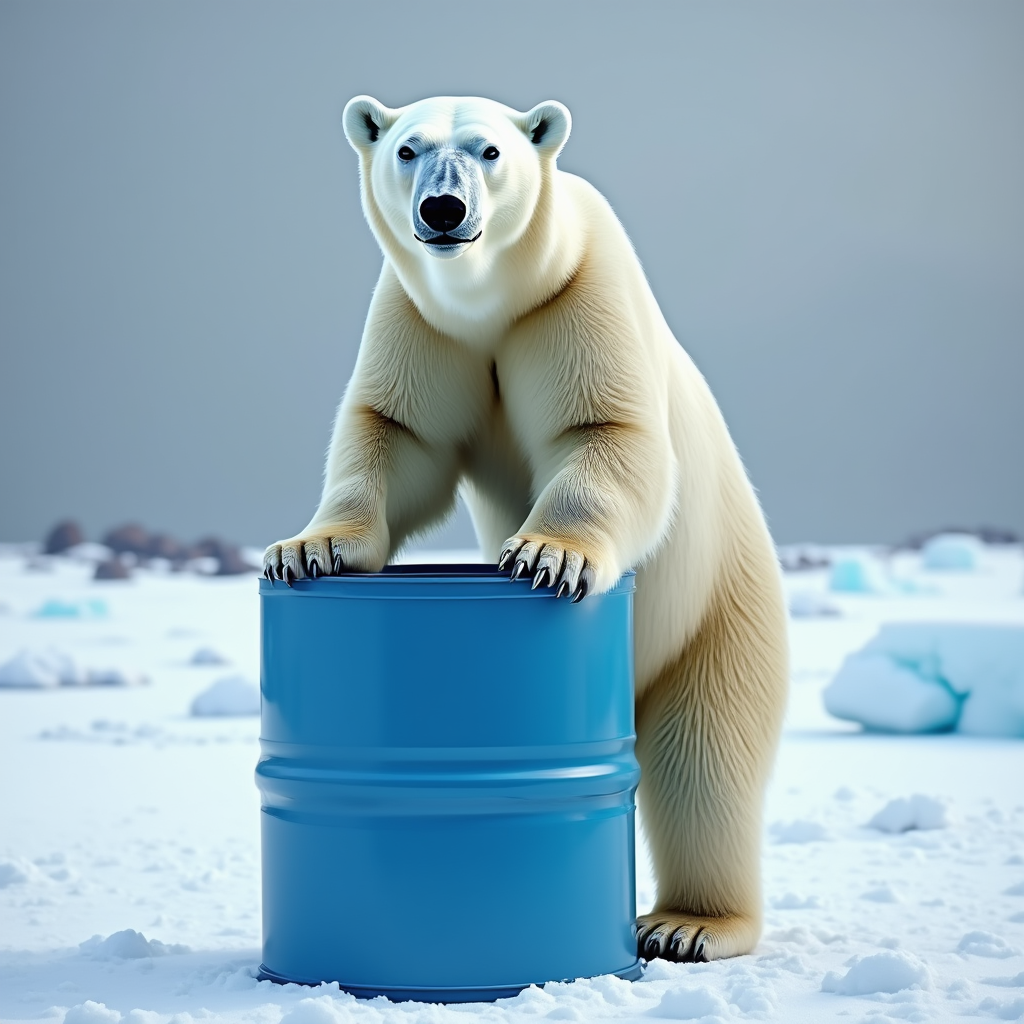} &
        \tightimg{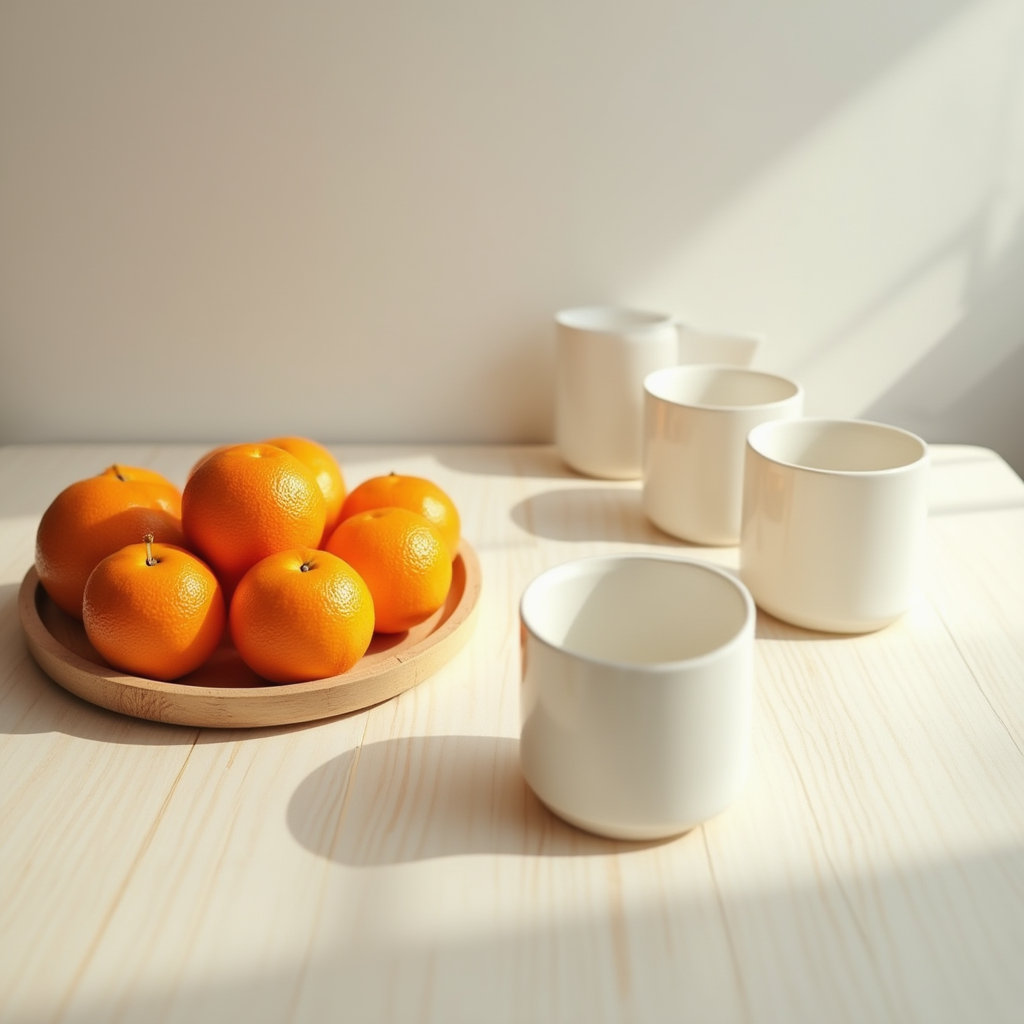} &
        \tightimg{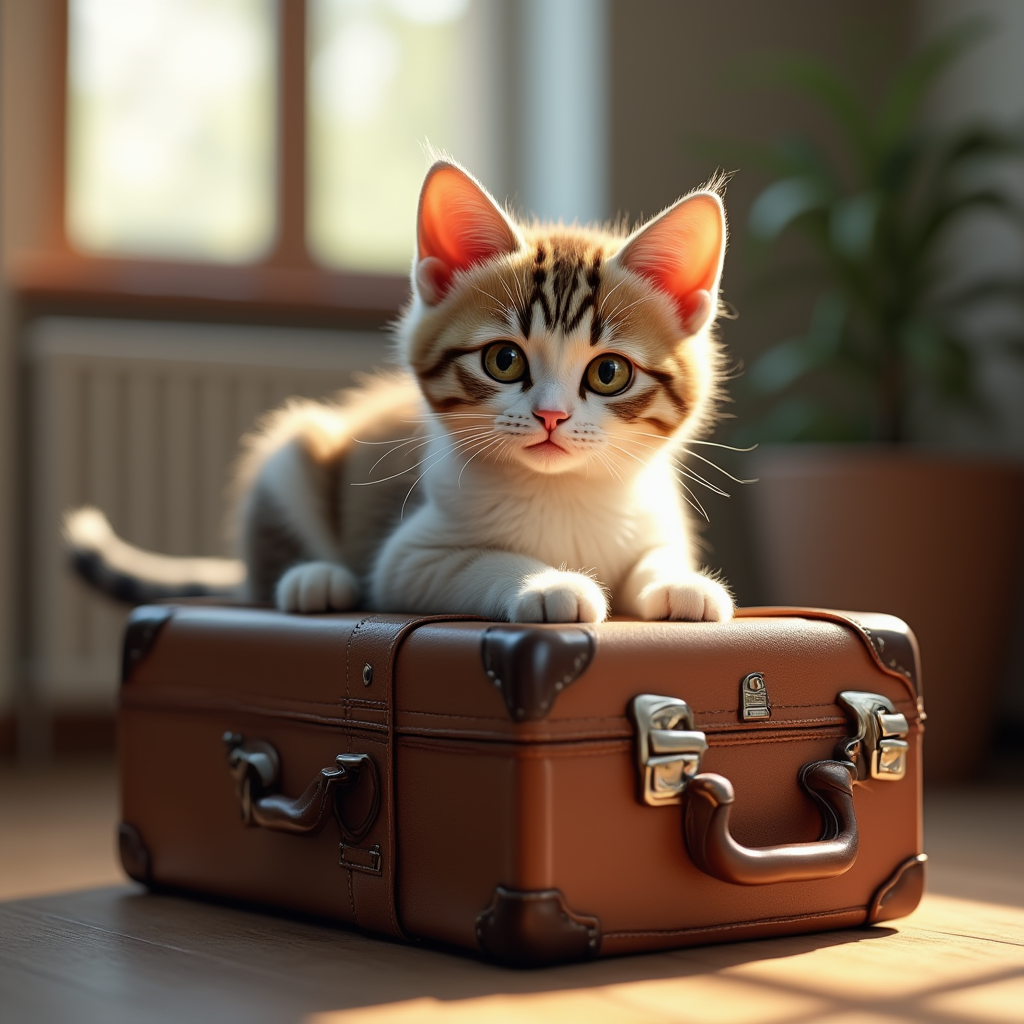} &
        \tightimg{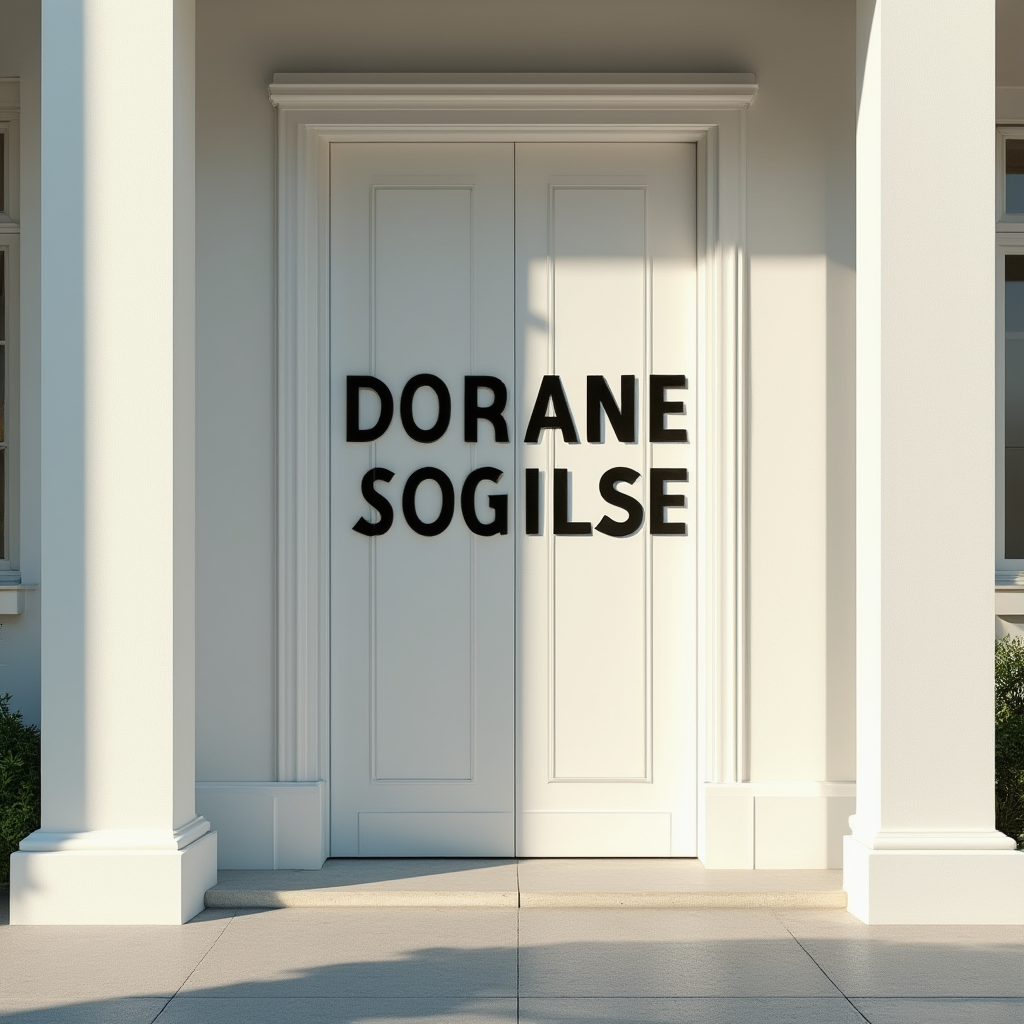} \\

        \begin{sideways}\small{\;\;\;\, MMaDA}\end{sideways} &
        \tightimg{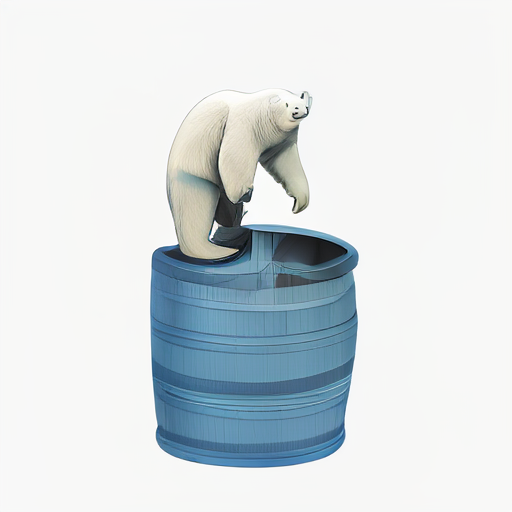} &
        \tightimg{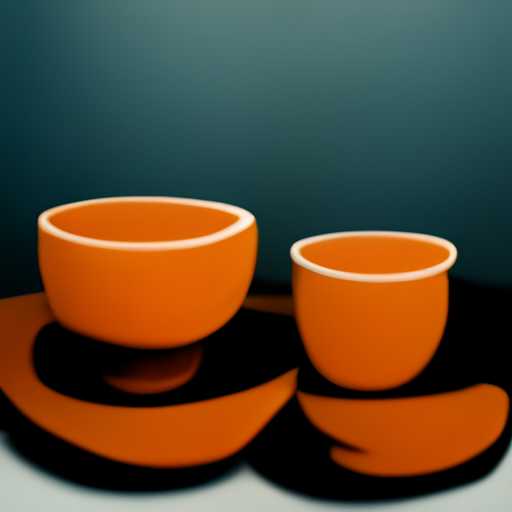} &
        \tightimg{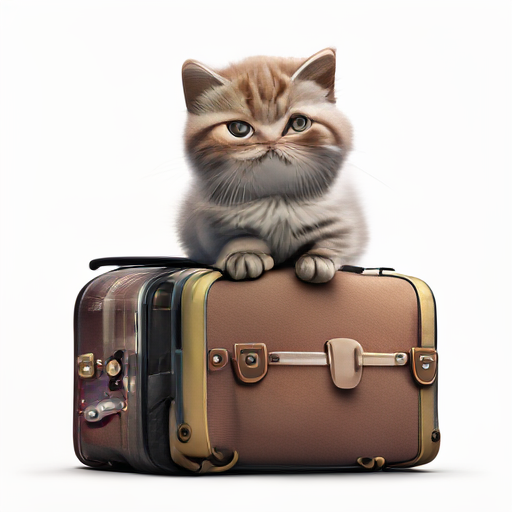} &
        \tightimg{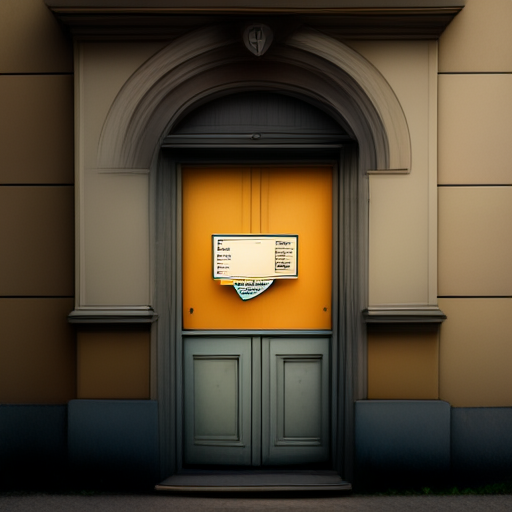} \\

        \begin{sideways}\small{\;\;\; \EF}\end{sideways} &
        \tightimg{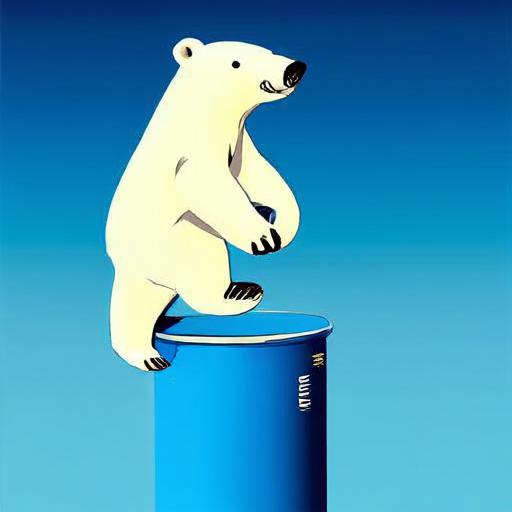} &
        \tightimg{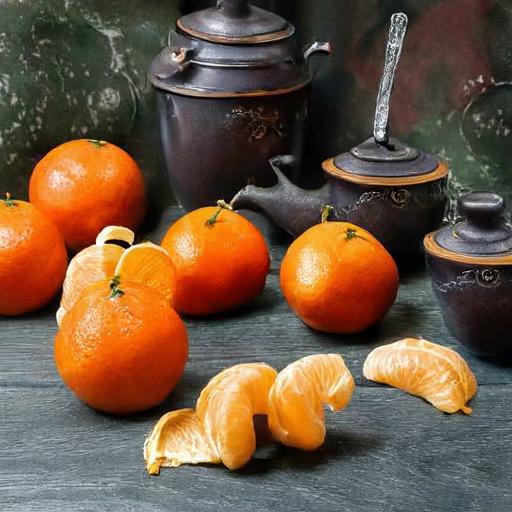} &
        \tightimg{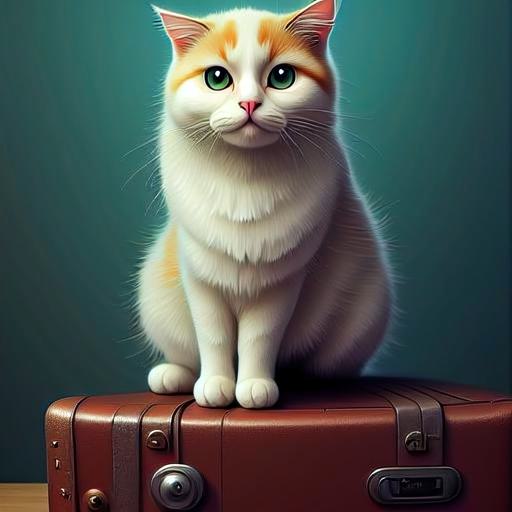} &
        \tightimg{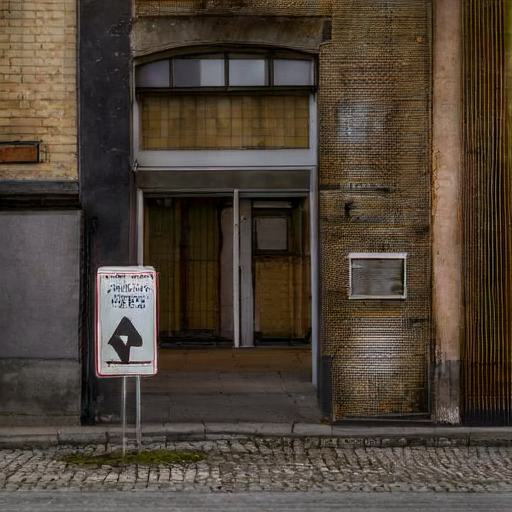}  
        \vspace{1em} \\
        &
        \parbox{23mm}{\centering\it A polar bear balancing on a blue barrel.} &
        \parbox{23mm}{\centering\it A table with some oranges and some cups.} &
        \parbox{23mm}{\centering\it A very cute cat sitting on a piece of luggage.} &
        \parbox{23mm}{\centering\it A building with a door sitting behind a sign.} \\
    \end{tabular}
    \end{adjustbox}
    \caption{\textbf{Qualitative comparison of \EF and SOTA models.} We notice that \EF gets the details of the prompt correctly, for instance the polar bear is \textit{`balancing on a blue barrel'}. The visual details of our generation are also better compared to MMaDA possibly due to using continuous image tokens rather than discrete. In the last column, the figure shows \EF handles common semantic challenges more effectively, as it was able to generate a building with \textit{`a door sitting behind a sign'}.}
    \vspace{-1pt}
    \label{fig:qualitative}
\end{figure*}

\subsection{Comparison with State-of-The-Art Unified Models}
\label{sec:comparison}
In Tables \ref{tab:pretraining_metrics} and \ref{tab:performance_comparison}, we benchmark \EF against other autoregressive and diffusion multimodal models. Our models deliver competitive performance across all tasks, rivaling state-of-the-art results despite significant differences in training procedures.
\paragraph{Understanding Performance.} On visual question answering, our 1B model nearly matches MMaDA 8B, while our 8B model surpasses MMaDA 8B on most benchmarks (Table \ref{tab:performance_comparison}). This is noteworthy because models like MMaDA received extensive post-training with chain-of-thought data and RL finetuning, whereas \EF did not. As discussed in Section \ref{sec:vqa_sampling}, we demonstrate that reasoning naturally emerges during generation. Similarly, while FUDOKI leveraged a pretrained multimodal model, \EF delivers competitive performance without this advantage. We observe that initializing from a pretrained text-only autoregressive model consistently enhances both prompt alignment for image generation and VQA performance, yielding similar gains as the AR+FM model (Table \ref{tab:llama_init}).
\paragraph{Generation Performance.} For image generation, we measured FID using open-source weights on COCO-2014~\citep{coco2014} validation set. Since MMaDA and Bagel received extensive aesthetic finetuning that produces images stylistically distant from real images (yielding higher FID scores). For fair comparison, we benchmark against MMaDA Base and Bagel without thinking mode. We show qualitative comparisons in Figure \ref{fig:qualitative}.

\subsection{OneFlow Unlocks New Capabilities}
\label{sec:new_capabilities}
\subsubsection{Classifier-free guidance improves text detailedness}

\begin{figure}[t]
    \centering
    \includegraphics[width=1.0\linewidth]{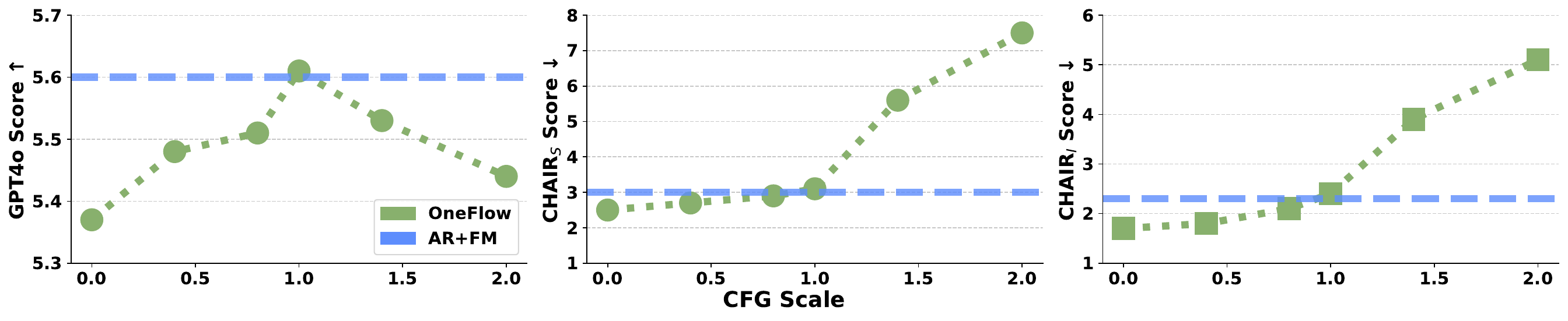}
    \caption{\textbf{\EF with classifier-free guidance produces longer and more detailed answers} \EF matches the AR baseline in GPT-4o as judge~\citep{caparena}, while higher guidance scales trade off between detailedness and hallucination. CHAIR$_S$ and CHAIR$_I$ measure the rate of hallucinated objects at the sentence and instance levels, respectively (lower is better).}
    \label{fig:cfg_hallucination}
    \vspace{-1em}
\end{figure}

\begin{figure}
    \small
    \setlength{\tabcolsep}{3pt}
    \centering
    \begin{adjustbox}{max width=1.0\textwidth}
    \begin{tabular}{@{} >{\centering\arraybackslash}p{0.27\linewidth} >{\raggedright\arraybackslash}p{0.72\linewidth} @{}}
    \toprule
    \vtop{\null\hbox{\includegraphics[width=\linewidth]{figures/images/000000065485.jpg}}} 
    \vspace{0.2em}
    Input image & %
    \vspace{-0.5em}%
    {\scriptsize\textbf{[CFG 0.0:]}} A dog sitting in the driver’s seat of a truck. 
    \vspace{0.2em} \newline 
    {\scriptsize\textbf{[CFG 1.0:]}} A dog sitting in the driver's seat of a \hlb{red and gray} truck. 
    \vspace{0.2em} \newline 
    {\scriptsize\textbf{[CFG 2.0:]}} A \hlb{red and grey suv} truck with a dog sitting in the driver's seat. \hlb{The truck is parked on a grassy field with a tree on the left side and a clear} \hlb{blue sky in the background.}
    \\
    \bottomrule
    \end{tabular}
    \end{adjustbox}%
    \caption{\textbf{Edit Flows with classifier-free guidance produces longer and more detailed answers}, examples of classifier-free guidance effects on text generation are shown in Figure~\ref{fig:image_caption_generation_more}.}
    \label{fig:image_caption_generation}
\end{figure}

The use of continuous-time Markov chains allows us to apply classifier-free guidance (CFG) to our model's insertion rates. 
Specifically, given an unconditional prediction $\lambda(X_t)Q(X_t)$ and a conditional prediction $\lambda(X_t | c)Q(X_t | c)$, where $c$ is the prompt and $w$ is the guidance weight, the modified insertion rate is:
\begin{equation}
    \lambda^\text{cfg}(X_t | c) = \lambda(X_t | c)^w \lambda(X_t)^{1-w} \quad \text{ and } \quad Q^\text{cfg}(X_t | c) \propto Q(X_t | c)^w Q(X_t)^{1-w}.
\end{equation}

As shown in Figures~\ref{fig:image_caption_generation} and \ref{fig:image_caption_generation_more}, higher CFG values consistently increase the length and detail of generated text. We quantitatively evaluated caption quality and hallucination using CapArena~\citep{caparena} by prompting GPT4-o. Figure~\ref{fig:cfg_hallucination} shows that increasing CFG leads to more detailed captions, with \EF matching \AR's level of detail at a guidance scale of 1. However, this increased detail comes at the expense of hallucinations at very high CFG values.

\subsubsection{Simultaneous generation of interleaved text and images}
\label{sec:interleaved}
When autoregressive multi-modal models insert an image, they append it at the end of the current generation, fully denoise it, then continue the generation process. However, OneFlow is able to simultaneously denoise images and the text. When the model deems it necessary, it is able to insert an new image in the existing text and denoise it along with the text, as proposed in Section \ref{sec:mixed_modal_schedules}.
%

To train this model, we took OneFlow 1B Mixed and finetuned it on the interleaved subset of the Chameleon dataset \citep{team2024chameleon} for $20000$ steps. This subset contains $17000$ examples that interleave both text and image data. 
Figure \ref{fig:color_coded_example} shows the generation order of the tokens where two images were generated as part of the answer, with more detailed examples in Appendix \ref{sec:further_interleaved} and animated versions in the supplementary material.

\section{Related Work}

\begin{figure}
    \centering
    {\large Text Generation}\vspace{1em} \\
    \begin{subfigure}[b]{0.33\linewidth}
    \centering
    \includegraphics[height=2.2cm]{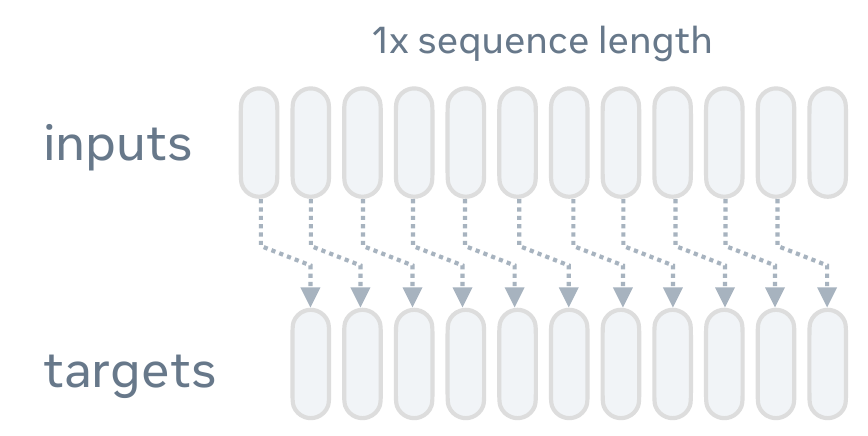}
    \caption*{\hspace{4em} Autoregressive}
    \end{subfigure}\hfill
    \begin{subfigure}[b]{0.25\linewidth}
    \centering
    \includegraphics[height=2.2cm]{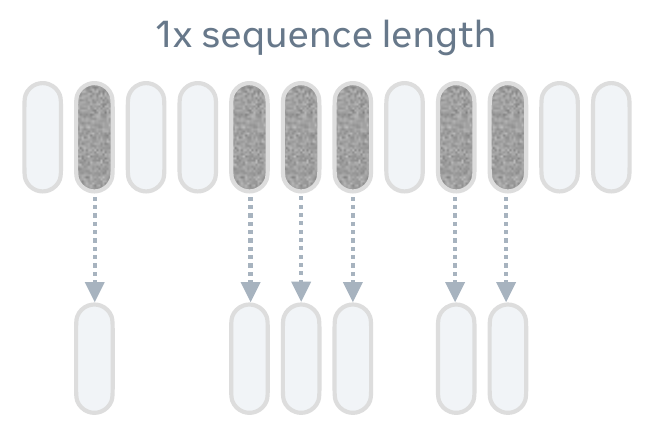}
    \caption*{Mask diffusion}
    \end{subfigure}\hfill
    \begin{subfigure}[b]{0.25\linewidth}
    \centering
    \includegraphics[height=2.1cm]{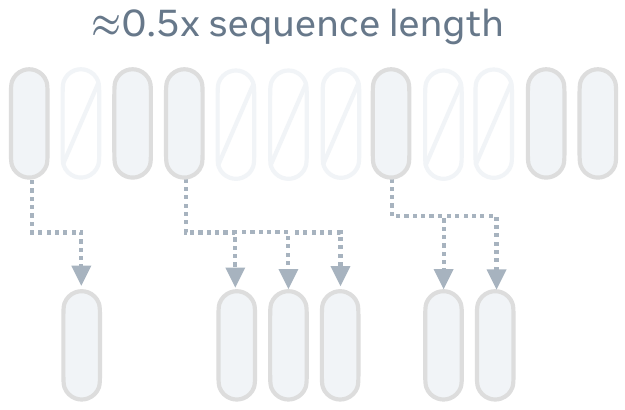}\vspace{0.3mm}
    \caption*{Edit Flow}
    \end{subfigure}\hfill
    \begin{subfigure}[b]{0.1\linewidth}
    \centering
    \includegraphics[height=1.4cm]{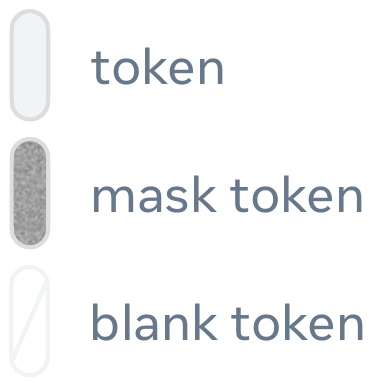}\vspace{9mm}
    \end{subfigure}\\
    \vspace{1em}
    {\large Interleaved Text and Image Generation}\vspace{1em} \\
    \begin{subfigure}[b]{0.39\linewidth}
    \centering
    \includegraphics[height=2.2cm]{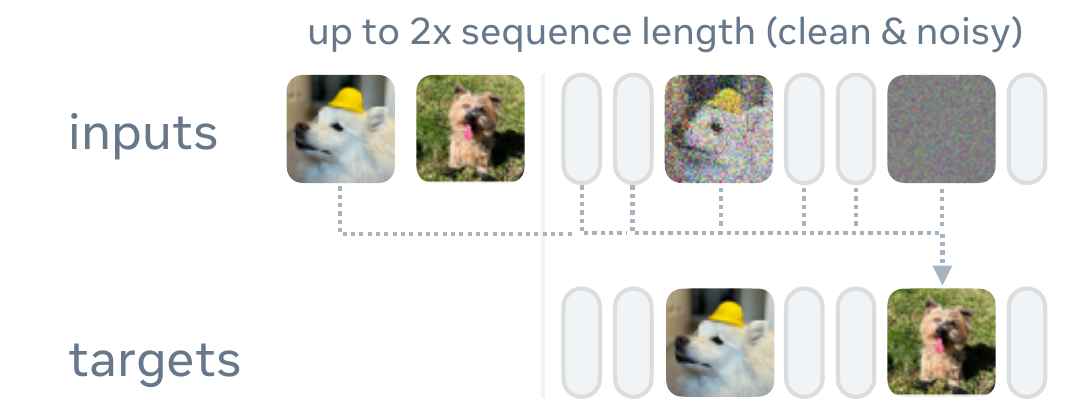}
    \caption*{\hspace{4em} Transfusion (autoreg. + diffusion)}
    \end{subfigure}\hfill
    \begin{subfigure}[b]{0.3\linewidth}
    \centering
    \includegraphics[height=2.2cm]{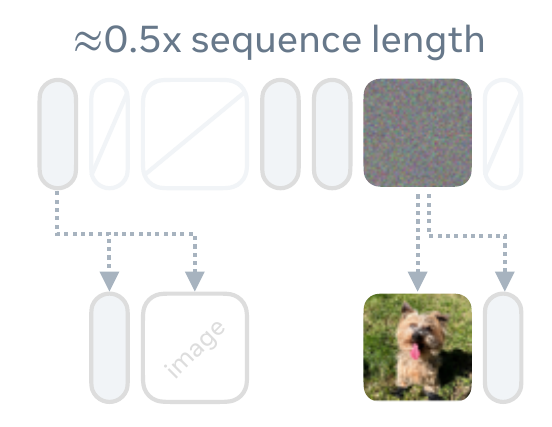}
    \caption*{OneFlow}
    \end{subfigure}\hfill
    \begin{subfigure}[b]{0.19\linewidth}
    \centering
    \includegraphics[height=1.6cm]{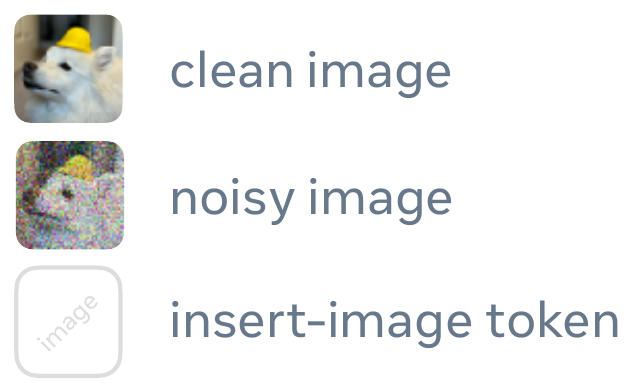}\vspace{8mm}
    \end{subfigure}
    \caption{Illustration of the model input and targets during training for (\textit{top}) text generation and (\textit{bottom}) interleaved generation. On text generation, Edit Flow simply deletes tokens instead of replacing them with a special mask token, resulting in lower FLOPS but with the same information as the mask diffusion framework.
    On interleaved generation, to train autoregressive with diffusion denoising, the images are typically duplicated so that both the clean and the noisy images are in the sequence. On the other hand, OneFlow deletes tokens and images during training which reduces the sequence length.}
    \label{fig:placeholder}
\end{figure}
\begin{table}
\centering
\renewcommand{\arraystretch}{1.1}
\begin{adjustbox}{max width=\textwidth, center}
\begin{tabular}{l c c c c c}
\toprule
\multicolumn{6}{l}{\; \textit{Generative Model Framework}} \\
Text & AR & AR & Mask Diffusion & Discrete FM & Edit Flow \\
Image & AR & Diffusion / FM & Mask Diffusion & Discrete FM & FM \\
\midrule
\multicolumn{6}{l}{\; \textit{Training Properties}} \\
Attention & Causal & Block Causal & Bidirectional & Bidirectional & Bidirectional \\
\rowcolor{metablue!10} Tokens / Iter & Seq Len & up to 2x Seq Len & Seq Len & Seq Len & $\sim$50\% Seq Len \\
\midrule
\multicolumn{6}{l}{\; \textit{Generation Capabilities}} \\
Understanding & \checkmark & \checkmark  & \checkmark  & \checkmark  & \checkmark  \\
Single image generation & \checkmark  & \checkmark  & \checkmark  & \checkmark  & \checkmark  \\
\rowcolor{metablue!10} Variable length & \checkmark & \checkmark  & $\times$  & $\times$  & \checkmark  \\
\rowcolor{metablue!10} Concurrent mixed-modal & $\times$ & $\times$ & \checkmark & \checkmark  & \checkmark  \\
\rowcolor{metablue!10} Interleaved generation & $\times$ & \checkmark  & $\times$ & $\times$ & \checkmark \\
\midrule
Examples & \begin{tabular}{@{}c@{}}Chameleon~\citep{team2024chameleon}\\JanusPro~\citep{januspro}\end{tabular} & \begin{tabular}{@{}c@{}}Transfusion~\citep{transfusion}\\Bagel~\citep{bagel}\end{tabular} & MMaDA~\citep{mmada} & FUDOKI~\citep{fudoki} & \textbf{\Large OneFlow} \\
\bottomrule
\end{tabular}
\end{adjustbox}
\caption{High-level comparison between different frameworks for building unified models of text and image generation.}
\label{tab:architecture_comparison}
\end{table}

\paragraph{Unified Multimodal Models.} 
Early unified models like Flamingo~\citep{alayrac2022flamingo} pioneered the integration of vision and language via cross-attention. Emu2~\citep{sun2023emu} advanced this paradigm by employing a unified autoregressive objective. Chameleon~\citep{team2024chameleon} adopted an early-fusion, AR architecture that unifies images and text as discrete tokens. Subsequent work spans three architectural paradigms: diffusion~\citep{mmada,ddit,unidisc,muddit,fudoki}, autoregressive~\citep{alayrac2022flamingo,team2024chameleon,sun2024emu2,januspro,janus,yu2023scalingautoregressivemultimodalmodels}, and hybrids~\citep{zhao2024monoformer,transfusion,showo,metamorph,janusflow, lmfusion}. Transfusion~\citep{transfusion} improves upon Chameleon by leveraging AR for text and diffusion for images. The Janus series collectively introduces decoupled visual pathways for understanding and generation~\citep{wu2024janus,ma2024janusflow,chen2025januspro}. BAGEL scales multimodal pretraining using a Mixture‑of‑Transformers (MoT)~\citep{liang2024mixture,deng2025bagel}. In parallel, diffusion‑based models have emerged as promising alternatives: DDiT applies discrete diffusion to text and continuous diffusion to images~\citep{ddit}; UniDisc uses fully discrete diffusion with Llama tokenizer for text and MAGVIT‑v2~\citep{yu2024languagemodelbeatsdiffusion} for images~\citep{unidisc}; FUDOKI uses discrete flow matching instead of masked diffusion, with a SigLIP~\citep{siglip} encoder for understanding and a LlamaGen~\citep{sun2024autoregressive} tokenizer for generation~\citep{fudoki}. MMaDA scales masked diffusion with mixed chain‑of‑thought fine‑tuning and UniGRPO~\citep{mmada}. MuDDiT combines a pretrained T2I (Meissonic~\citep{bai2024meissonic}) with a lightweight text decoder for MaskGIT‑style discrete diffusion~\citep{chang2022maskgit,muddit}. While these models are limited by a fixed generation order or fixed-length outputs, our approach fundamentally differs by simultaneously generating interleaved content and a variable number of images. We summarize the differences between each framework in Table \ref{tab:architecture_comparison}.

\paragraph{Discrete Diffusion and Discrete Flow Matching.} 
Iterative refinement models, including diffusion \citep{sohl2015deep,ho2020denoising} and flow models \citep{liu2022flow,albergo2023stochastic,lipman2024flow}, have been adapted for discrete token spaces. Discrete diffusion models typically learn to reverse a corruption process \citep{austin2021structured,lou2024discrete}, while discrete flow models transport between two distributions with an interpolating scheme \citep{campbell2024generative,gat2024discrete}. 
These frameworks offer a large design space for discrete generative modeling \citep{shaul2024flow, fudoki}, and can further be generalized to a large family of continuous-time Markov processes with a unified training recipe \citep{holderrieth2024generator}. However, recent works on discrete diffusion have predominantly focused on a simplified mask construction \citep{sahoo2024simple,shi2024simplified,ou2024your,zheng2024masked}. This masking framework, however, cannot be easily applied to variable-length and especially simultaneous interleaved generation. 

\paragraph{Edit-based Non-autoregressive Language Models.} 
Early non-autoregressive models for variable-length generation \citep{gu2019insertion,gu2019levenshtein,stern2019insertion,reid2022diffuser} often relied on multiple models and evaluations to handle edit operations, \eg by inserting placeholder tokens first and then replacing them with the actual tokens. While later work like Edit Flows \citep{havasi2025edit} improved on this by using a continuous-time framework and removes the need for any placeholder tokens. In order to finetune from existing mask diffusion models, \citet{Dreamon2025} and \citet{kim2025any} have also explored inserting only mask tokens, breaking down insertions back into two steps. In contrast, we simply made use of autoregressive models as initialization, modifying only model outputs for insertion predictions with no special tokens used in the model inputs, as this allows us to save on training FLOPS and inference costs. Related to combining discrete and continuous generative processes, 
\citet{campbell2024trans} also proposed modeling insertions with a diffusion model for denoising, but only considered singular insertions and required an additional generative model due to not using an interleaved time schedule.
In contrast, our approach considers mixed-modal sequence data, denoising images concurrently with per-image time values, and allows for parallel token insertions.

\section{Conclusion and Limitations}

We introduced \EF, a novel non-autoregressive multimodal model that overcomes the fixed-length generation limitations of diffusion models and has better scaling than autoregressive multimodal models. 
We introduced mixed-modal generation approaches, which through extensive controlled experiments,  improve on benchmarks for both image understanding and image generation. 
We also propose a novel approach to interleaved generation that simultaneously denoises images and inserts text tokens, with promising qualitative results. Interleaved generation is still in its infancy and we expect to see more incoming research efforts in constructing large-scale data sets~\citep{mint,obelics,mmc4} and designing comprehensive benchmarks.

A limitation of requiring bidirectional attention is the lack of key-value caching, which increases inference cost. We do find that OneFlow can already obtain good captioning performance with very few model evaluations---outperforming AR with only 6 sampling steps (Figure \ref{fig:sampling_steps}). Therefore, further reducing inference costs, with semi-autoregressive models \citep{arriola2025block,gat2025set} or more sophisticated methods, would be an exciting research direction.

\section*{Acknowledgements}

We thank Maha Elbayad, Emily Dinan, Xiaochuang Han, Lili Yu, Chunting Zhou, and Melissa Hall for building the Transfusion code base which we built our code upon. We thank Peter Tong and David Fan for pointers on VQA and feedback on the paper. We thank Koustuv Sinha, Sharut Gupta, Marjan Ghazvininejad, Jakob Verbeek, and Amir Bar for feedback on the paper draft and fruitful discussions throughout the project.

\bibliographystyle{assets/plainnat}
\bibliography{paper}

\appendix

\section{Additional Generation Examples} \label{sec:further_interleaved} 

\begin{figure}[H]
    \centering
    \setlength{\fboxsep}{0pt}
    \setlength{\fboxrule}{0.6pt}

    \begin{subfigure}[t]{0.22\textwidth}
        \fbox{\includegraphics[width=\linewidth]{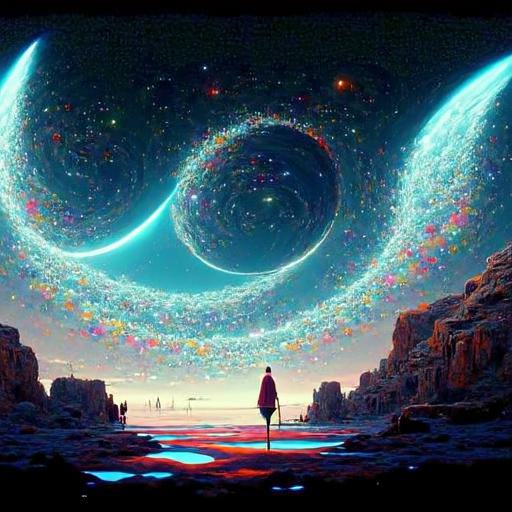}}
        \caption*{\centering \textit{A man standing in the middle of a nebula filled with stars.}}
    \end{subfigure}\hfill
    \begin{subfigure}[t]{0.22\textwidth}
        \fbox{\includegraphics[width=\linewidth]{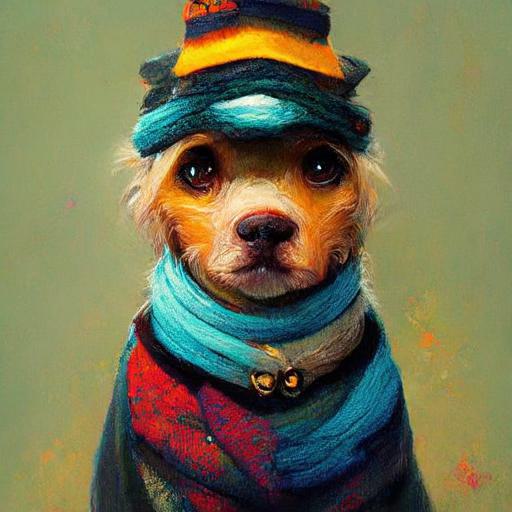}}
        \caption*{\centering \textit{A dog wearing a hat and a scarf.}}
    \end{subfigure}\hfill
    \begin{subfigure}[t]{0.22\textwidth}
        \fbox{\includegraphics[width=\linewidth]{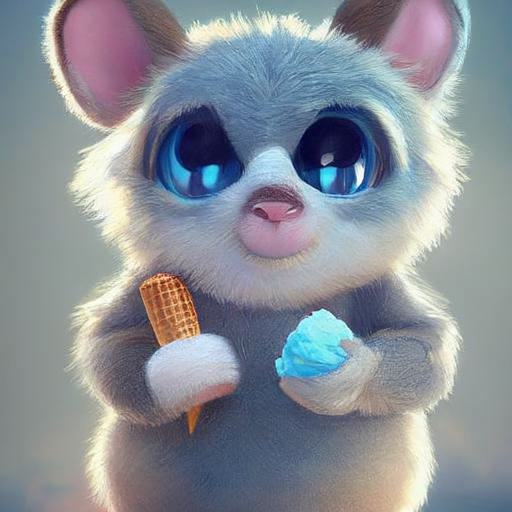}}
        \caption*{\centering \textit{A cute animal holding an ice cream cone.}}
    \end{subfigure}\hfill
    \begin{subfigure}[t]{0.22\textwidth}
        \fbox{\includegraphics[width=\linewidth]{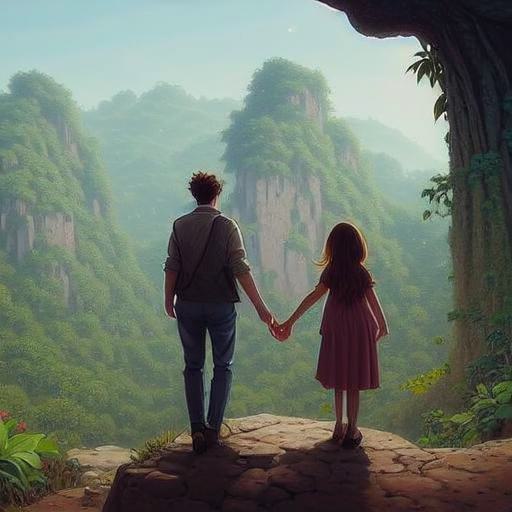}}
        \caption*{\centering \textit{Lovely adventurer and explorer in tropical forest with mountains.}}
    \end{subfigure}

    \begin{subfigure}[t]{0.22\textwidth}
        \fbox{\includegraphics[width=\linewidth]{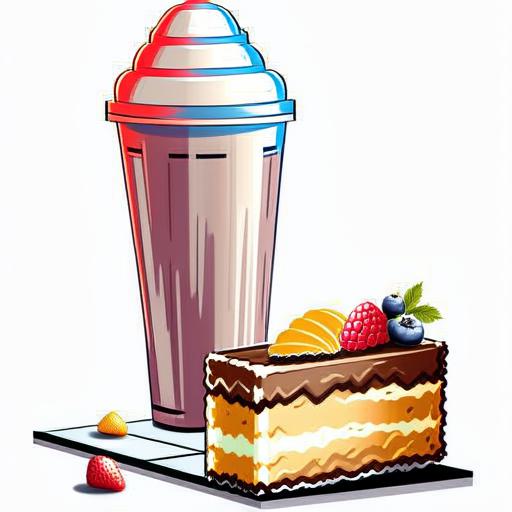}}
        \caption*{\centering \textit{A shake next to a cake.}}
    \end{subfigure}\hfill
    \begin{subfigure}[t]{0.22\textwidth}
        \fbox{\includegraphics[width=\linewidth]{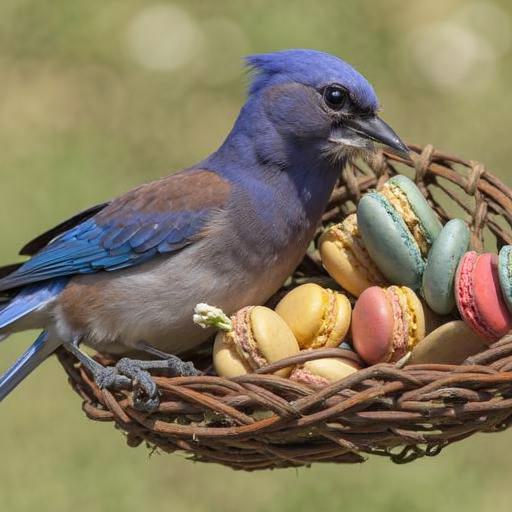}}
        \caption*{\centering \textit{A blue jay standing on a large basket of rainbow macarons.}}
    \end{subfigure}\hfill
    \begin{subfigure}[t]{0.22\textwidth}
        \fbox{\includegraphics[width=\linewidth]{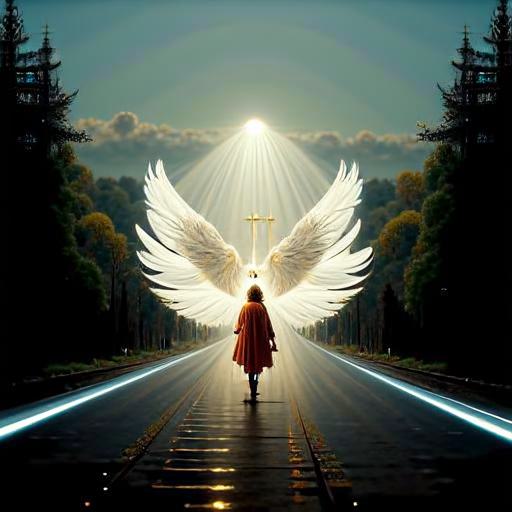}}
        \caption*{\centering \textit{An angel in the middle of the road.}}
    \end{subfigure}\hfill
    \begin{subfigure}[t]{0.22\textwidth}
        \fbox{\includegraphics[width=\linewidth]{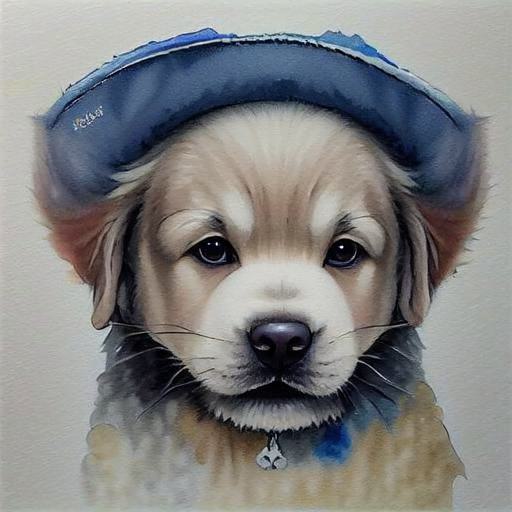}}
        \caption*{\centering \textit{Cute and fluffy Labrador puppy in hat. Watercolor painting.}}
    \end{subfigure}

    \begin{subfigure}[t]{0.22\textwidth}
        \fbox{\includegraphics[width=\linewidth]{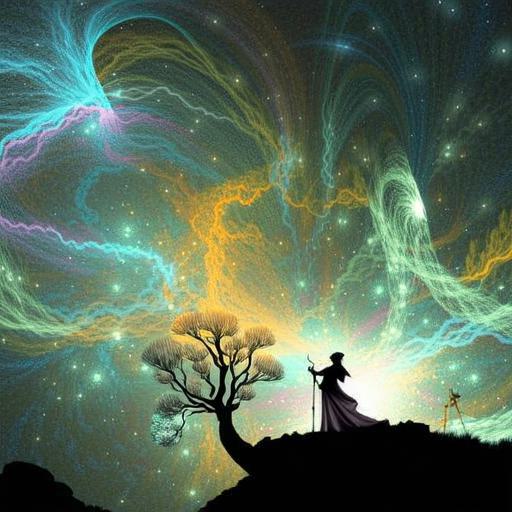}}
        \caption*{\centering \textit{Space scene. Magician standing on landscape silhouette with tree and fractal colorful nebula.}}
    \end{subfigure}\hfill
    \begin{subfigure}[t]{0.22\textwidth}
        \fbox{\includegraphics[width=\linewidth]{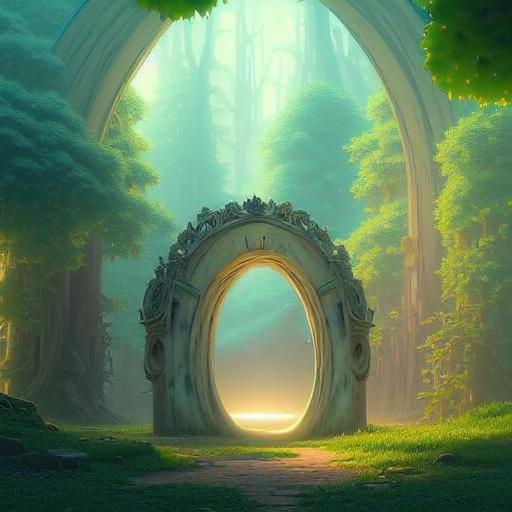}}
        \caption*{\centering \textit{Fantasy portal in forest landscape surrounded by tree and magic. Green nature with magical architecture concept art.}}
    \end{subfigure}\hfill
    \begin{subfigure}[t]{0.22\textwidth}
        \fbox{\includegraphics[width=\linewidth]{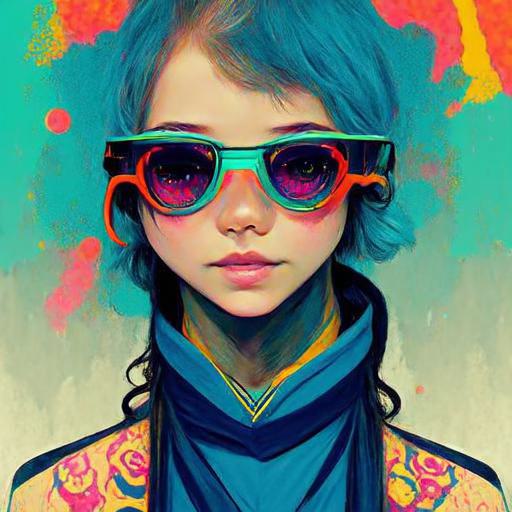}}
        \caption*{\centering \textit{Portrait of a hipster wearing sunglasses. Stylized drawing of a fashionista on a colored background.}}
    \end{subfigure}\hfill
    \begin{subfigure}[t]{0.22\textwidth}
        \fbox{\includegraphics[width=\linewidth]{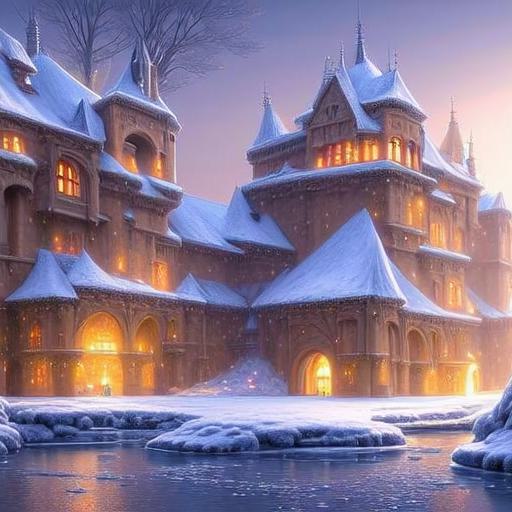}}
        \caption*{\centering \textit{A majestic castle illuminated by lights. The ground is covered in a blanket of snow, with a body of water in the foreground.}}
    \end{subfigure}

    \begin{subfigure}[t]{0.22\textwidth}
        \fbox{\includegraphics[width=\linewidth]{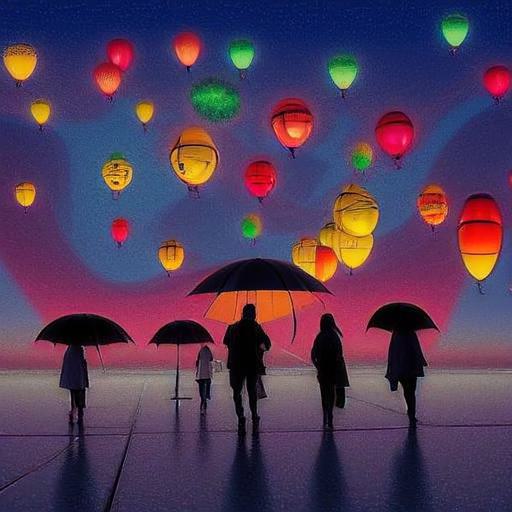}}
        \caption*{\centering \textit{People walking with umbrellas with balloons in the sky}}
    \end{subfigure}\hfill
    \begin{subfigure}[t]{0.22\textwidth}
        \fbox{\includegraphics[width=\linewidth]{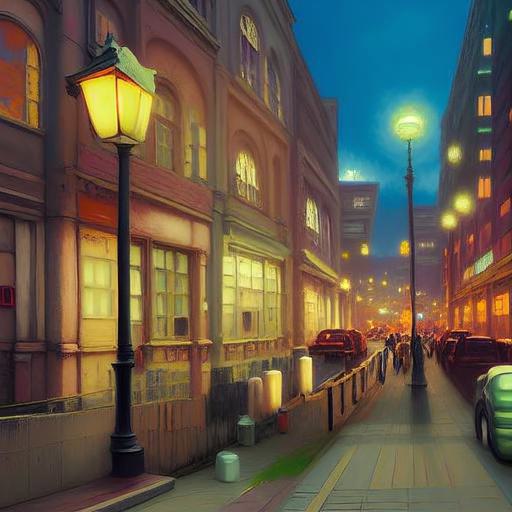}}
        \caption*{\centering \textit{A painting of a city street at night.}}
    \end{subfigure}\hfill
    \begin{subfigure}[t]{0.22\textwidth}
        \fbox{\includegraphics[width=\linewidth]{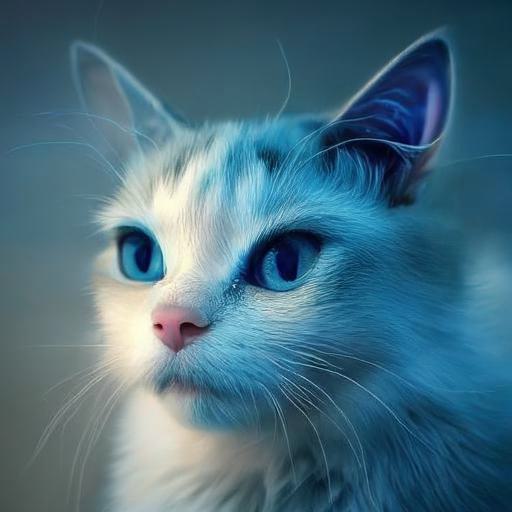}}
        \caption*{\centering \textit{A blue cat.}}
    \end{subfigure}\hfill
    \begin{subfigure}[t]{0.22\textwidth}
        \fbox{\includegraphics[width=\linewidth]{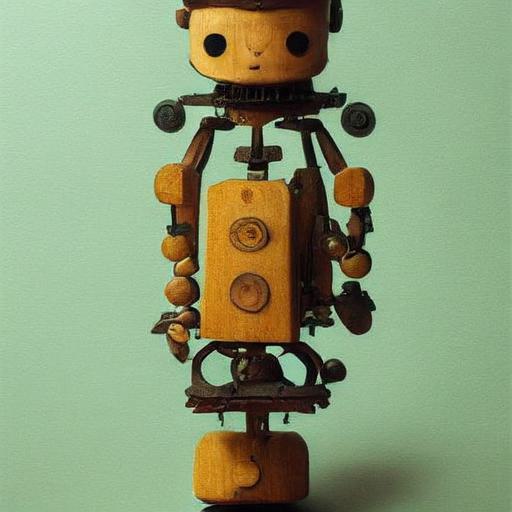}}
        \caption*{\centering \textit{A robot made out of wood.}}
    \end{subfigure}

    \caption{Generated images from the OneFlow model at 512$\times$512 resolution.}
    \label{fig:generated_images_with_prompts}
\end{figure}

\begin{figure}[H]
\centering
\begin{tabular}{m{0.75\textwidth}m{0.24\textwidth}}
\textbf{Generated Text} & \textbf{Generated Images} \\
\hline
\textsc{\textbf{Prompt:}} I'm looking for a new hobby that's relaxing and will allow me to be outdoors to get some fresh air. Offer me a suggestion and show me an image of the type of place I can do it

& \\
{\small\textsc{$\mathbf{t=0.2}$:}}  
? It's that. senseness tranqu promotes. 
 & No image \\
\\
{\small\textsc{$\mathbf{t=0.4}$:}} 
Why try? It's a that you a wide of, lake trail. in can create senseness tranqu promotes stress and.
 & No image \\
\\
{\small\textsc{$\mathbf{t=0.6}$:}}  
Why not try nature? It's a that you a wide variety of natural, like this lake trail,. in can create sense ofness and tranqu promotes stress and.
 & No image \\
\\
{\small\textsc{$\mathbf{t=0.8}$:}} 
Why not try nature? It's a relaxing that allows you to in a wide variety of natural environments, like this lake trail, for instance.\texttt{\color{red}<|image|>}Walking in can create sense of calmness and tranqu that promotes stress and relaxation.
 & \includegraphics[width=2.5cm,height=2.5cm,keepaspectratio]{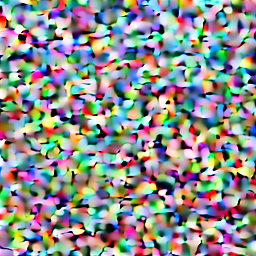} \\
\\

{\small\textsc{$\mathbf{t=1.0}$:}} 
Why not try nature walking? It's a relaxing hobby that allows you to immerse yourself in a wide variety of natural environments, like this lake trail, for instance.\texttt{\color{red}<|image|>}Walking in nature can create a sense of calmness and tranquility that promotes stress reduction and relaxation.
 & \includegraphics[width=2.5cm,height=2.5cm,keepaspectratio]{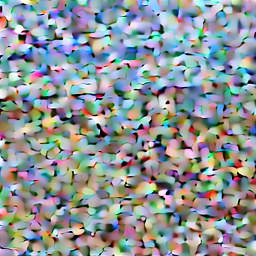}   \\
\\
{\small\textsc{$\mathbf{t=2.0}$:}} 
Why not try nature walking? It's a relaxing hobby that allows you to immerse yourself in a wide variety of natural environments, like this lake trail, for instance.\texttt{\color{red}<|image|>}Walking in nature can create a sense of calmness and tranquility that promotes stress reduction and relaxation.
 &
\includegraphics[width=2.5cm,height=2.5cm,keepaspectratio]{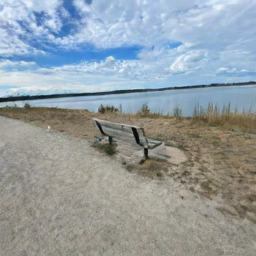} \\
\\

\end{tabular}
\caption{Example interleaved generation process showing outdoor exercise. In this example, the image is inserted late in the generation process when the text is almost fully denoised.} \label{fig:interleaved_generation_2}
\end{figure}

\begin{figure}[H]
\centering
\resizebox{!}{0.5\textheight}{
\begin{tabular}{m{0.75\textwidth}m{0.24\textwidth}}
\textbf{Generated Text} & \textbf{Generated Images} \\
\hline
{\small \textsc{\textbf{Prompt:}}} I want to paint flowers but I need a photo for inspiration. Can you show me two pictures of flowers and describe them?

& \\
{\small \textsc{$\mathbf{t=0.2}$:}}  first a pink of out in appealing roses and overall the. seems be the, appreciation for.

 a pink a flower pink with possibly vase is base the flower appeal simple backdrop the point of the image
& No image \\
\\
{\small\textsc{$\mathbf{t=0.4}$:}} In first, a arrangement pink flowers is, of six flowers out in appealing manner.\texttt{\color{red}<|image|>} roses are delicate and pink the overall of the. The bouquet seems be the captured,oking and appreciation for the nature.

This image a view pink sitting a table flower be ger with a pink with prominent center possibly a. vase is base the flower visual appeal table a simple backdrop, allowing flower be the point of the image.

 & \includegraphics[width=2.0cm,height=2.0cm,keepaspectratio]{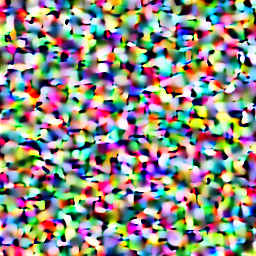} \\
\\
{\small \textsc{$\mathbf{t=0.6}$:}}
In first, a arrangement of pink flowers is, of six flowers out in a appealing manner.\texttt{\color{red}<|image|>} roses are delicate and have a, pink, the overall of the. The bouquet seems be focus the image captured, showcasing their The display and atmosphere,oking of and appreciation for the nature.

This image a view of a pink sitting a table.\texttt{\color{red}<|image|>} flower appears be ger with a pink with prominent center possibly a bud. vase is the base the flower, enhancing visual appeal of the. The table vase a simple backdrop, allowing flower be the point of the image.

 & \includegraphics[width=2.0cm,height=2.0cm,keepaspectratio]{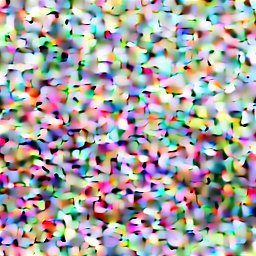}  \includegraphics[width=2.0cm,height=2.0cm,keepaspectratio]{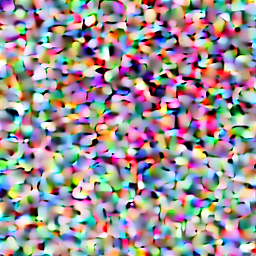}  \\
\\
{\small\textsc{$\mathbf{t=0.8}$:}} In this first image, a beautiful arrangement of pink flowers is on display, with total of six flowers spread out in a appealing manner.\texttt{\color{red}<|image|>}The roses are delicate and have a soft, pink hue, which adds the overall charm of the scene. The bouquet seems be the main focus of the image flowers captured a close shot, showcasing their intricate details. The display and atmosphere, evoking a sense of romance and appreciation for the beauty nature.

This image a view of a pink flower sitting a glass a table.\texttt{\color{red}<|image|>}The flower appears be ger with a pink color with a a prominent center, possibly a bud. vase is the base of the flower, enhancing visual appeal of the. The table the vase provides a simple backdrop, allowing flower to be the focal point of the image.

 & \includegraphics[width=2.0cm,height=2.0cm,keepaspectratio]{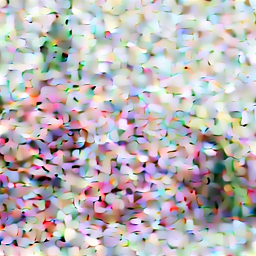} 
  \includegraphics[width=2.0cm,height=2.0cm,keepaspectratio]{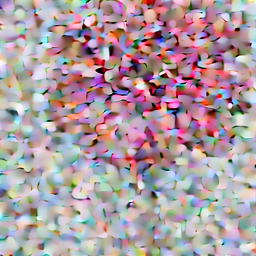} \\
\\

{\small\textsc{$\mathbf{t=1.0}$:}} In this first image, a beautiful arrangement of pink flowers is on display, with a total of six flowers spread out in a visually appealing manner.\texttt{\color{red}<|image|>}The roses are delicate and have a soft, pink hue, which adds to the overall charm of the scene. The bouquet seems to be the main focus of the image, with the flowers captured in a close-up shot, showcasing their intricate details. The display creates an elegant and serene atmosphere, evoking a sense of romance and appreciation for the beauty of nature.

This image features a close-up view of a beautiful pink flower sitting in a glass vase on a table.\texttt{\color{red}<|image|>}The flower appears to be a gerbera, with a vibrant pink color, with a vibrant pink color and a prominent center, possibly a bud. The glass vase is positioned at the base of the flower, enhancing the visual appeal of the scene. The table beneath the vase provides a simple yet elegant backdrop, allowing the flower to be the focal point of the image.

 &  \includegraphics[width=2.0cm,height=2.0cm,keepaspectratio]{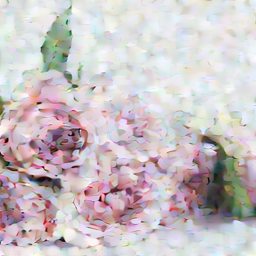} \includegraphics[width=2.0cm,height=2.0cm,keepaspectratio]{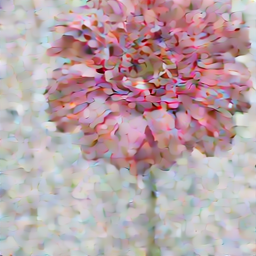}  \\
\\

{\small\textsc{$\mathbf{t=2.0}$:}} In this first image, a beautiful arrangement of pink flowers is on display, with a total of six flowers spread out in a visually appealing manner.\texttt{\color{red}<|image|>}The roses are delicate and have a soft, pink hue, which adds to the overall charm of the scene. The bouquet seems to be the main focus of the image, with the flowers captured in a close-up shot, showcasing their intricate details. The display creates an elegant and serene atmosphere, evoking a sense of romance and appreciation for the beauty of nature.

This image features a close-up view of a beautiful pink flower sitting in a glass vase on a table.\texttt{\color{red}<|image|>}The flower appears to be a gerbera, with a vibrant pink color, with a vibrant pink color and a prominent center, possibly a bud. The glass vase is positioned at the base of the flower, enhancing the visual appeal of the scene. The table beneath the vase provides a simple yet elegant backdrop, allowing the flower to be the focal point of the image.

 & \includegraphics[width=2.0cm,height=2.0cm,keepaspectratio]{figures/interleaved_figures/flowers/decoded_run_11_step_2999_t_1.9999_batch_0.png}  \includegraphics[width=2.0cm,height=2.0cm,keepaspectratio]{figures/interleaved_figures/flowers/decoded_run_11_step_2999_t_1.9999_batch_1.png}  \\
\end{tabular}
}
\vspace{-1em}
\caption{Example interleaved generation process showing flowers.} \label{fig:interleaved_generation}
\end{figure}

\begin{figure}[H]
\centering
\resizebox{!}{0.5\textheight}{
\begin{tabular}{m{0.75\textwidth}m{0.24\textwidth}}
\textbf{Generated Text} & \textbf{Generated Images} \\
\hline
\textsc{\textbf{Prompt:}} I am teaching an ESL lesson on home types. Please show me images of a bungalow, a two-story home and a townhouse that I can show my students, and briefly explain about each of them.

& \\
{\small\textsc{$\mathbf{t=0.2}$:}}    good. **B is typically for with a-p simple including suburban towns. is of a.Two**
 type residential building levels with the typically bedrooms on the. example-story. **Town town unit town. They vary typically than homes image example.\texttt{\color{red}<|image|>}

 & \includegraphics[width=2.0cm,height=2.0cm,keepaspectratio]{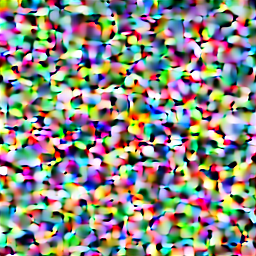}  \\
\\
{\small\textsc{$\mathbf{t=0.4}$:}}    which good show to them about different types1. **Bungalow**
ungalow is of is typically cosy designed for. Bungal are single-story, with a-p simple can including suburban neighbourhood towns and areas. Below is of a.Two-**
A a type residential building levels, with the typically on bedrooms on the. below example of-story. **Town A town multi residential unit that town walls. They can vary size are typically smaller than single homes image is example of.\texttt{\color{red}<|image|>}

 & 
 \includegraphics[width=2.0cm,height=2.0cm,keepaspectratio]{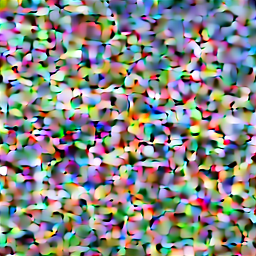}
 \\
\\
{\small\textsc{$\mathbf{t=0.6}$:}}   images which would good to show students to them about different types:

1. **Bungalow**
ungalow is a type of is typically small, cosy and designed for comfortable. Bungal are often single-story, with a-p simple can found various locations including suburban neighbourhood towns and areas. Below is an image of a.Two-**
A two a type residential building has levels, with the typically on the first floor and bedrooms and additional on the. image below an example of-story. **Town A town multi-level residential unit that attached town shared walls. They can vary size, are typically smaller than single homes. The image is example of of.\texttt{\color{red}<|image|>}

 & 
 \includegraphics[width=2.0cm,height=2.0cm,keepaspectratio]{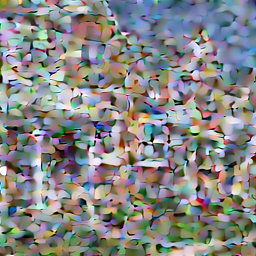}
 \\
\\
{\small\textsc{$\mathbf{t=0.8}$:}} are images which would be good to show your ESL students to them about different types:

1. **Bungalow**
A bungalow is a type of that is typically small, cosy and designed for comfortable living. Bungalows are often single-story, with a-pitched and a simple floor can be found various locations, including suburban neighbourhoods, beach towns and areas. Below is an image of a b2.Two-Story home**
A two home a type residential building has levels or floors, with the typically on the first floor and bedrooms and additional spaces on the second. image below is an example of a two-story home.3. **Townhouse**
 A town a multi-level residential unit that attached one more other town shared walls. They can vary size, but are typically smaller than single-family homes. The image below is example of a ofhouses.\texttt{\color{red}<|image|>}

 & 
 \includegraphics[width=2.0cm,height=2.0cm,keepaspectratio]{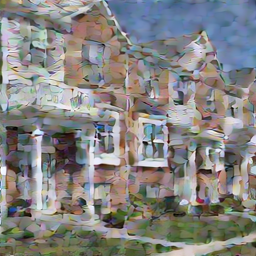}
 \\
\\

{\small\textsc{$\mathbf{t=1.0}$:}} Here are three images which would be good to show your ESL students to teach them about different home types:

1. **Bungalow**
A bungalow is a type of house that is typically small, cosy, and designed for comfortable living. Bungalows are often single-story homes, with a low-pitched roof and a simple, open floor plan. They can be found in various locations, including suburban neighbourhoods, beach towns, and rural areas. Below is an image of a bungalow:\texttt{\color{red}<|image|>}2. **Two-Story home**
A two-story home is a type of residential building that has two levels or floors, with the main living areas typically on the first floor and bedrooms and additional living spaces on the second floor. The image below is an example of a two-story home.\texttt{\color{red}<|image|>}3. **Townhouse**
 A townhouse is a multi-level residential unit that is attached to one or more other townhouses via shared walls. They can vary in size, but are typically smaller than single-family homes. The image below is an example of a row of townhouses.\texttt{\color{red}<|image|>}

 &

 \includegraphics[width=2.0cm,height=2.0cm,keepaspectratio]{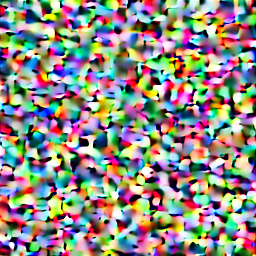}
 \includegraphics[width=2.0cm,height=2.0cm,keepaspectratio]{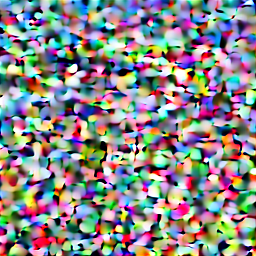}
 \includegraphics[width=2.0cm,height=2.0cm,keepaspectratio]{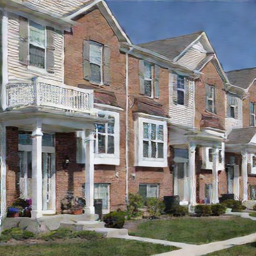}
 \\
\\

{\small\textsc{$\mathbf{t=2.0}$:}} Here are three images which would be good to show your ESL students to teach them about different home types:

1. **Bungalow**
A bungalow is a type of house that is typically small, cosy, and designed for comfortable living. Bungalows are often single-story homes, with a low-pitched roof and a simple, open floor plan. They can be found in various locations, including suburban neighbourhoods, beach towns, and rural areas. Below is an image of a bungalow:\texttt{\color{red}<|image|>}2. **Two-Story home**
A two-story home is a type of residential building that has two levels or floors, with the main living areas typically on the first floor and bedrooms and additional living spaces on the second floor. The image below is an example of a two-story home.\texttt{\color{red}<|image|>}3. **Townhouse**
 A townhouse is a multi-level residential unit that is attached to one or more other townhouses via shared walls. They can vary in size, but are typically smaller than single-family homes. The image below is an example of a row of townhouses.\texttt{\color{red}<|image|>}

 &

 \includegraphics[width=2.0cm,height=2.0cm,keepaspectratio]{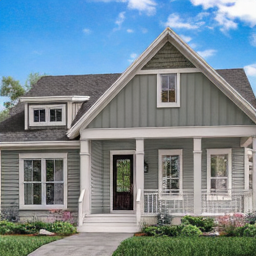}
 \includegraphics[width=2.0cm,height=2.0cm,keepaspectratio]{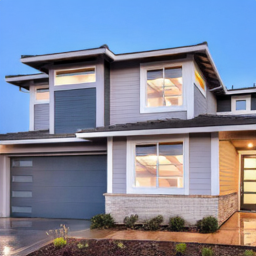}
 \includegraphics[width=2.0cm,height=2.0cm,keepaspectratio]{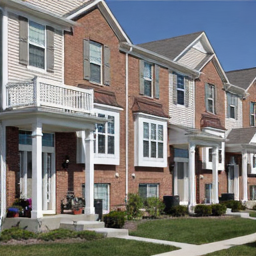}
 \\
\end{tabular}
}
\vspace{-1em}
\caption{Example interleaved generation process showing home types. An animated version is included in the supplementary material.} \label{fig:interleaved_generation_3}
\end{figure}

\begin{figure}[H]
    \vspace{-3em} 
    \scriptsize
    \setlength{\tabcolsep}{1pt}
    \centering
    \begin{adjustbox}{max width=\textwidth}
    \begin{tabular}{@{} >{\centering\arraybackslash}p{0.27\linewidth} >{\raggedright\arraybackslash}p{0.72\linewidth} @{}}
    \toprule
    \textbf{Input image} & \textbf{Captions with varying classifier-free guidance weights} \\
    \midrule
    \vspace{-0.5em}
    \vtop{\null\hbox{\includegraphics[width=\linewidth]{figures/images/000000252216.jpg}}} & %
    \vspace{1em}%
    {\scriptsize\textbf{[CFG 0.0:]}} A llama and a horse standing in a field.
    \vspace{1em} \newline 
    {\scriptsize\textbf{[CFG 2.0:]}} A \hlb{white} llama and a \hlb{white} horse standing in a field.
    \\
    \midrule
    \vspace{-0.5em}
    \vtop{\null\hbox{\includegraphics[width=\linewidth]{figures/images/000000289393.jpg}}} & 
    {\scriptsize\textbf{[CFG 0.0:]}} A group of toy animals sitting on a table. 
    \vspace{0.2em} \newline 
    {\scriptsize\textbf{[CFG 1.0:]}} A table topped with a variety of toy animals, including a \hlb{giraffe}, a \hlb{cow}, and a \hlb{bird}, as well as a \hlp{toy duck}. In the background, \hlb{there is a frame attached to the wall.} 
    \vspace{0.2em} \newline 
    {\scriptsize\textbf{[CFG 1.5:]}} A \hlb{green} table topped with toy animals, including a giraffe, a cow, a \hlb{yellow} bird, and a toy duck, next to a \hlb{red box}. In the background, there is a \hlp{photo} frame attached to the wall.
    \\
    \midrule
    \vspace{-0.5em}
    \vtop{\null\hbox{\includegraphics[width=\linewidth]{figures/images/000000442009.jpg}}} & %
    \vspace{0.8em}
    {\scriptsize\textbf{[CFG 0.0:]}} A bathroom with a large tub and a sink. \vspace{0.2em} \newline 
    {\scriptsize\textbf{[CFG 1.0:]}} A bathroom with a \hlb{claw foot tub} and \hlb{three windows}. \vspace{0.2em} \newline 
    {\scriptsize\textbf{[CFG 2.5:]}} A \hlb{brown bathroom or master bathroom} with a classic claw foot tub and three windows. 
    \\
    \midrule
    \vspace{-0.5em}
    \vtop{\null\hbox{\includegraphics[width=\linewidth]{figures/images/000000049810.jpg}}} & %
    \vspace{0.5em}
    {\scriptsize\textbf{[CFG 0.0:]}} A cat sitting on a wooden deck looking up. \vspace{0.2em} \newline 
    {\scriptsize\textbf{[CFG 1.0:]}} A cat sitting on a wooden deck looking at its \hlb{reflection} in a window. \vspace{0.2em} \newline 
    {\scriptsize\textbf{[CFG 2.5:]}} \hlp{Two} fluffy \hlb{ginger and white} cats sit and gaze at their reflection in a \hlb{glass} window on a \hlb{green} wooden deck in \hlp{Japan}. 
    \\
    \midrule
    \vspace{-0.5em}
    \vtop{\null\hbox{\includegraphics[width=\linewidth]{figures/images/000000182417.jpg}}} & %
    \vspace{0.5em}
    {\scriptsize\textbf{[CFG 0.0:]}} A white plate topped with a cake and a spoon. \vspace{0.2em} \newline 
    {\scriptsize\textbf{[CFG 1.0:]}} A plate with a \hlb{dessert} and \hlb{two} spoons on it. \vspace{0.2em} \newline 
    {\scriptsize\textbf{[CFG 2.0:]}} A white plate topped with \hlb{ice cream}, accompanied by two spoons, a \hlp{bottle}, a \hlb{glass}, and a \hlb{tissue paper} on the \hlb{table}. \hlb{Through the glass window in the background, we can see the water and the sky.} 
    \\
    \midrule
    \vspace{-0.5em}
    \vtop{\null\hbox{\includegraphics[width=\linewidth]{figures/images/000000285349.jpg}}} & %
    \vspace{1em}%
    {\scriptsize\textbf{[CFG 0.0:]}} A glass bowl filled with colorful paper cranes.
    \vspace{0.2em} \newline 
    {\scriptsize\textbf{[CFG 1.0:]}} Colorful \hlb{origami} cranes in a glass bowl \hlb{shaped like a heart}.
    \vspace{0.2em} \newline 
    {\scriptsize\textbf{[CFG 2.0:]}} A table with a heart-shaped bowl filled with colorful origami cranes \hlb{in various colors}. \hlb{The background is slightly blurred}, giving the focus to the vibrant colors of the cranes.
    \\
    \bottomrule
    \end{tabular}
    \end{adjustbox}
    \caption{Text generation examples from OneFlow, which allows the use of classifier-free guidance (CFG). We observe that CFG produces longer and more \hlb{detailed} captions and also increased chance of \hlp{hallucinations}. Highlighted text show increased levels of detail when using higher CFG weights.}
    \label{fig:image_caption_generation_more}
\end{figure}

\section{Additional Experiment Results}
\subsection{VQA Average Performance Across Categories}
\label{sec:vqa_avg}
\begin{figure}[H]
    \centering
    \includegraphics[width=\linewidth]{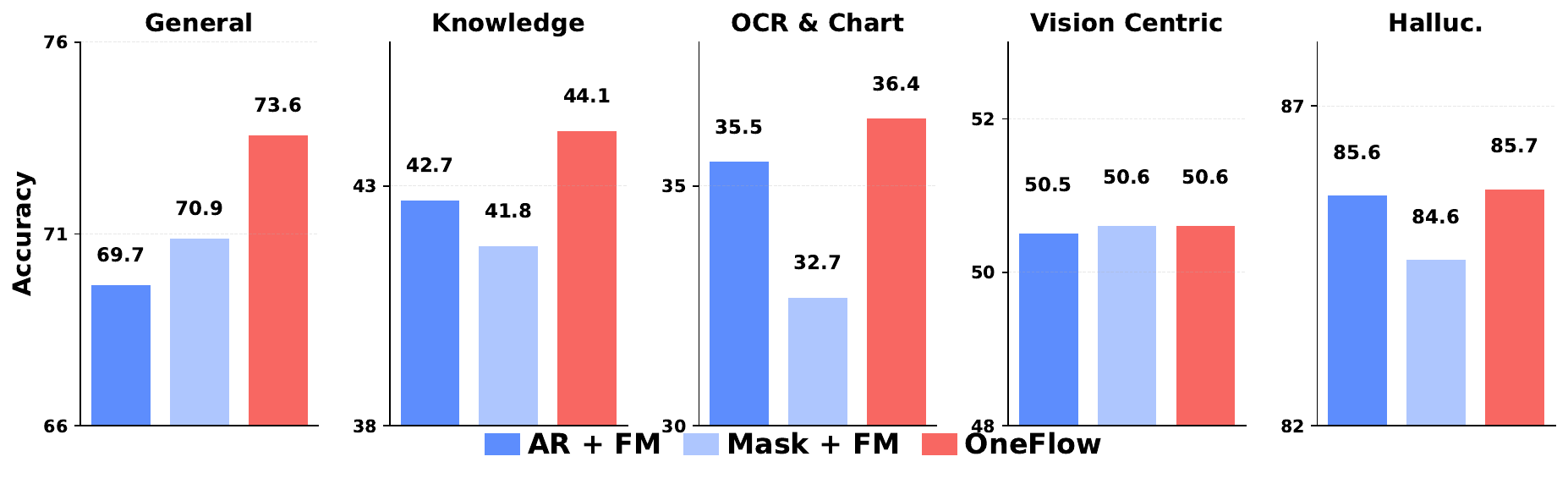}
    \caption{\textbf{VQA performance across categories in controlled experiments.} The largest gaps appear in General and OCR/Chart tasks.The models are tied on Vision-Centric benchmarks, which primarily depend on the visual encoder rather than text generation.}
    \label{tab:vqa_controlled_barplot}
\end{figure}

\subsection{Performance Between \AR and \EF During Pretraining}
\label{sec:pretraining}
\begin{figure}[H]
    \centering
    \includegraphics[width=\linewidth]{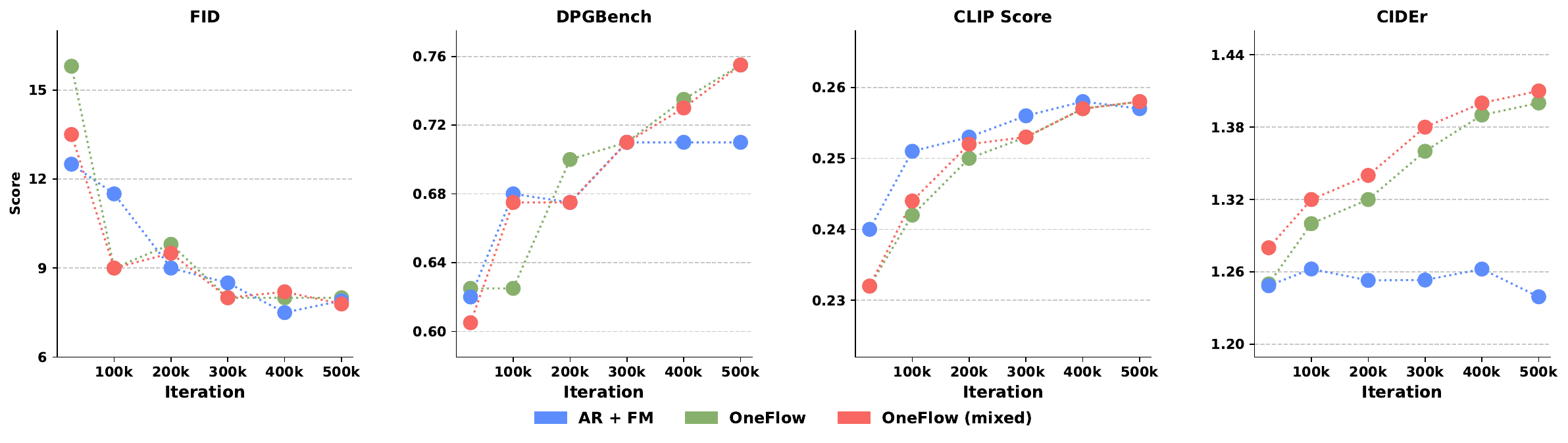}
    \caption{\textbf{Training curve for \EF vs. \AR for multimodal pretraining.} \EF initially starts out lower than \AR however it quickly catches up and exceeds \AR, most notably on DPG and CIDEr.}
    \label{fig:multimodal_pretraining}
\end{figure}

\subsection{Pretraining from Scratch vs LLama Init}
\label{sec:ablation_init}
\begin{table}[H]
\renewcommand{\arraystretch}{1.05}
\centering
\begin{adjustbox}{max width=\textwidth}
\begin{tabular}{@{} l c cccc c}
\toprule
& & \multicolumn{4}{c}{\cellcolor{colGen}\textbf{Image Generation}} & \cellcolor{colHalluc}\textbf{VQA} \\
\cmidrule(lr){2-5} \cmidrule(l){6-6}
\textbf{Model} & \textbf{Initialization} &
\cellcolor{colGen}\textbf{DPG$\uparrow$} & 
\cellcolor{colGen}\textbf{FID$\downarrow$} & 
\cellcolor{colGen}\textbf{CLIP$\uparrow$} & 
\cellcolor{colGen}\textbf{CIDEr$\uparrow$} & 
\cellcolor{colHalluc}\textbf{Avg VQA$\uparrow$} \\
\midrule
\EF & Random & 73.17 \hphantom{{\tiny($+0.00$)}} & 7.96 \hphantom{{\tiny($+0.00$)}} & 25.7 \hphantom{{\tiny($+0.0$)}} & 139.4 \hphantom{{\tiny($+0.0$)}} & 51.2 \hphantom{{\tiny($+0.0$)}} \\
\EF & LLaMA & 75.41 {\color{metablue}\tiny($+2.24$)} & 7.79 {\color{metablue}\tiny($-0.17$)} & 26.0 {\color{metablue}\tiny($+0.3$)} & 138.2 {\color{red}\tiny($-1.2$)} & 52.2 {\color{metablue}\tiny($+1.0$)} \\
\EF Mixed & Random & 74.86  \hphantom{{\tiny($+0.00$)}} & 7.69 \hphantom{{\tiny($+0.00$)}} & 25.8 \hphantom{{\tiny($+0.0$)}} & 140.0 \hphantom{{\tiny($+0.0$)}} & 51.6 \hphantom{{\tiny($+0.0$)}} \\
\EF Mixed & LLaMA & 75.08 {\color{metablue}\tiny($+0.22$)} & 7.44 {\color{metablue}\tiny($-0.25$)} & 25.8 {\tiny($+0.0$)} & 139.1 {\color{red}\tiny($-0.9$)} & 52.8 {\color{metablue}\tiny($+1.2$)} \\
\midrule
AR + FM & Random & 71.90 \hphantom{{\tiny($+0.00$)}} & 7.83 \hphantom{{\tiny($+0.00$)}} & 25.8 \hphantom{{\tiny($+0.0$)}} & 122.9 \hphantom{{\tiny($+0.0$)}} & 46.6 \hphantom{{\tiny($+0.0$)}} \\
AR + FM & LLaMA & 73.40 {\color{metablue}\tiny($+1.50$)} & 7.91 {\color{metablue}\tiny($-0.08$)} & 25.7 {\color{red}\tiny($-0.1$)} & 123.9 {\color{metablue}\tiny($+1.0$)} & 49.0 {\color{metablue}\tiny($+2.4$)} \\
\bottomrule
\end{tabular}
\end{adjustbox}
\caption{\textbf{Ablation study comparing LLaMA initialization vs. random initialization.} Except for CIDEr, using LLaMA as initialization generally offers benefits, especially for dense prompt image generation (DPG) and for VQA performance. Image generation metrics use CFG=3, and VQA results are averaged across benchmarks.}
\label{tab:llama_init}
\end{table}

\subsection{Sampling Steps on Caption Performance}
\label{sec:sampling step}

\begin{figure}[H]
\centering
\includegraphics[width=\textwidth]{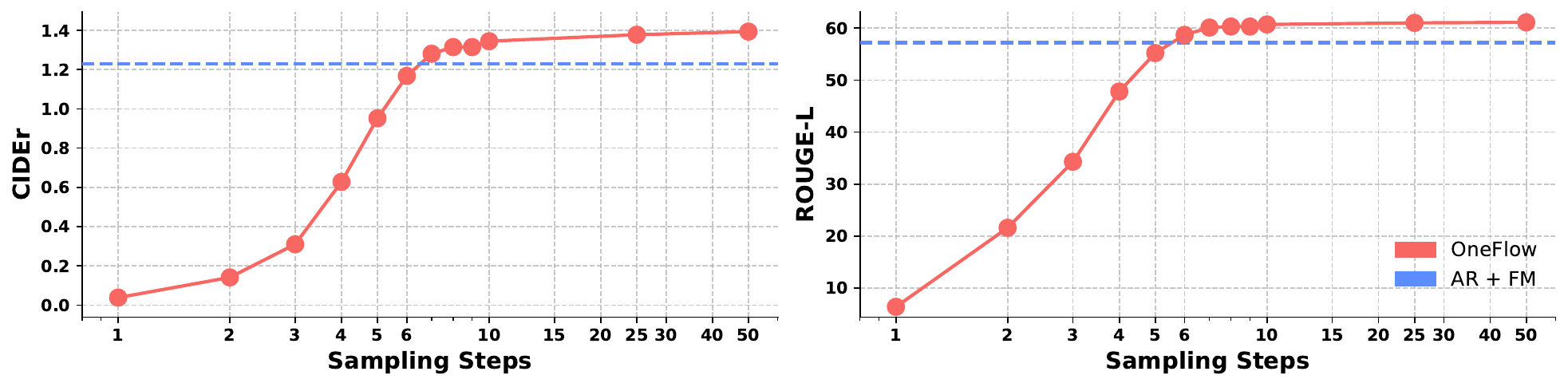}
\caption{\textbf{Performance vs. sampling steps compared to \AR.} OneFlow achieves parity with the AR model using only 6 sampling steps.}
\label{fig:sampling_steps}
\end{figure}

\subsection{t-Independent Parameterization}
\label{sec:t-independent}
\begin{table}[h]
\centering
\begin{tabular}{@{}cc@{}}
\toprule
\textbf{Time Location} & \textbf{CIDEr} \\
\midrule
\textbf{none} & \textbf{1.28} \\
begin\_of\_seq & 1.27 \\
begin\_of\_text & 0.33 \\
\bottomrule
\end{tabular}
\caption{CIDEr scores for different time embedding locations.}
\label{tab:time_location}
\end{table}
We empirically found t-independent parameterization to perform better despite theoretical motivation for t-dependence. This likely occurs because $X_t$ already encodes sufficient information about the noise level. We ablated the location of time token insertion: \texttt{begin\_of\_seq} places it at the beginning of the sequence, \texttt{begin\_of\_text} at the start of text tokens, and \texttt{none} corresponds to t-independent parameterization.

\subsection{K-scheduler}
\label{sec:kscheduler}

\begin{figure}[H]
    \centering
    \includegraphics[width=\linewidth]{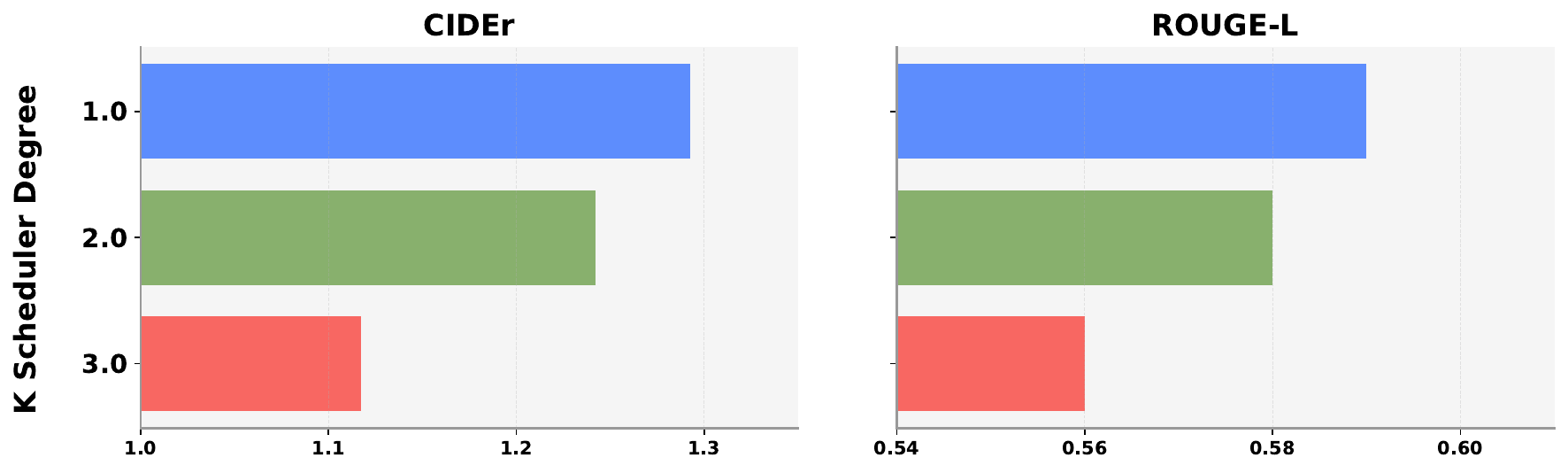}
    \caption{\textbf{Effect of $k$-scheduler degree.} Using the schedule $\kappa_t = t^k$, we find that linear scheduling ($k=1.0$) achieves the best results. Higher degrees lead to overly aggressive token deletion.}
    \label{fig:kscheduler}
    \vspace{-1em}
\end{figure}

We conducted ablations over different k-schedulers. As described in~\citep{gat2024discrete}, there are three variants of the k-scheduler: linear, quadratic, and cubic. We find that the linear scheduler works best.

\section{\EF Architecture}
\begin{figure}[H]
    \centering
    \includegraphics[width=0.8\linewidth]{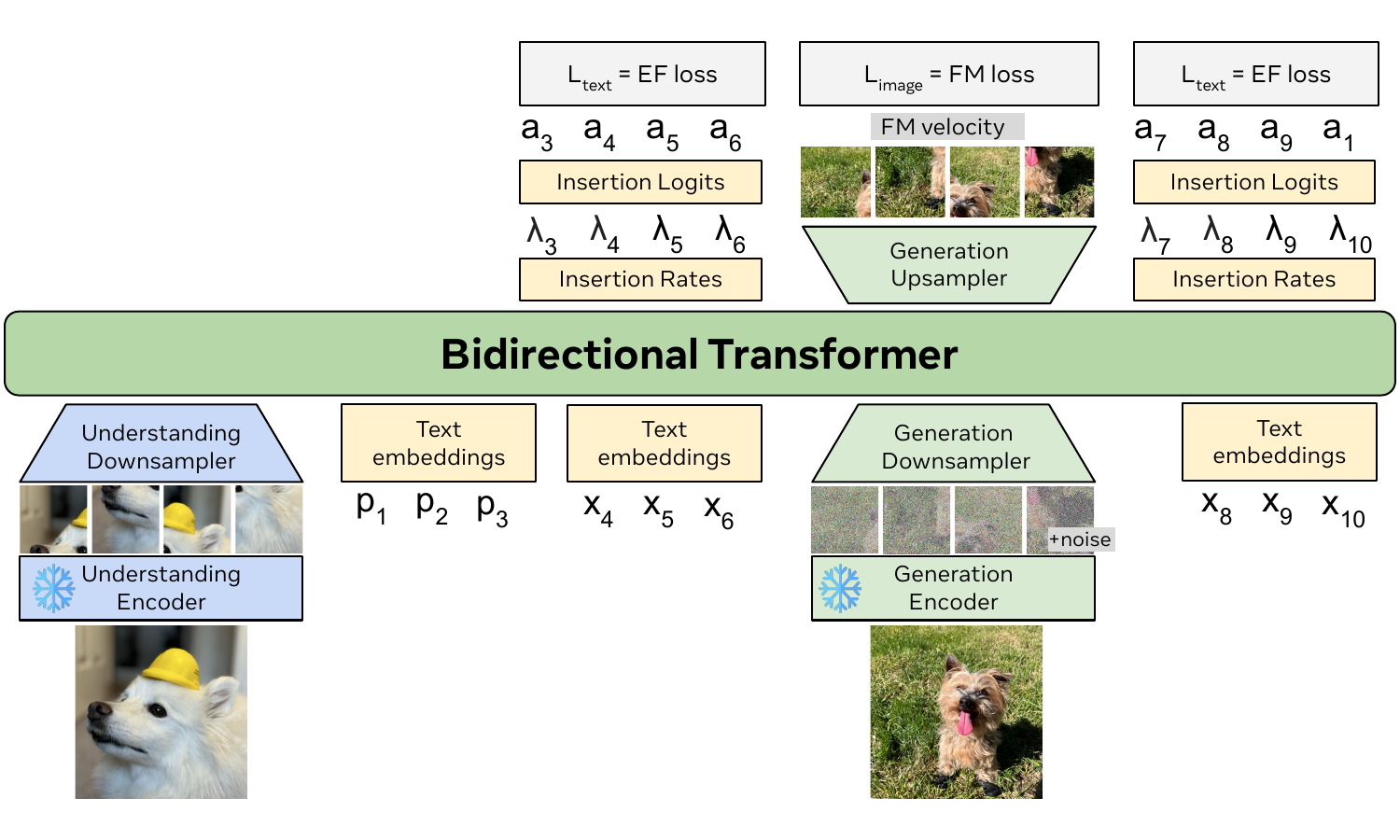}
\caption{\textbf{Architecture.} With a multimodal prompt, OneFlow can produce variable length generations with interleaved text \& images in a unified non-autoregressive sequence model, simultaneously generating all modalities with an interleaved time schedule for each generated image and text.}
    \label{fig:architecture}
\end{figure}

\section{Full Derivations}\label{app:edit_flow_derivation}

We provide the derivations of the model here. We briefly summarize the Edit Flow \citep{havasi2025edit} formulation and derivation, and then derive the interleaved time schedule when insertions and image denoising are performed simultaneously. 

\paragraph{Setup.} We make use of a blank token $\blank$ to denote empty spaces within a sequence. This token is only used for tracking token deletions during training and is not part of the vocabulary. Let $\mathcal{Z} = \bigcup_{n=0}^N (\gT \cup \{\blank\})^n$ be an extended space of aligned sequences. Furthermore, define $\strip: \mathcal{Z} \rightarrow \mathcal{X}$ as the function that removes all blank tokens from the sequence. Lastly, we define the delta function over sequences $\delta_{z_1}(z_2) = \prod_i \delta_{z_1^i}(z_2^i)$ which is one if all tokens are the same otherwise zero (i.e. Kronecker's delta function).

\paragraph{Continuous-time Markov chain (CTMC).} A CTMC is a continuous-time discrete-space process which iteratively jumps between discrete values, with transitions
\begin{equation}
    \P(X_{t+h} | X_t) = \delta_{X_t}(X_{t+h}) + h u_t(x | X_t) + o(h),
\end{equation}
where $u_t$ can be interpreted as a first-order characterization of the transition kernel.
Since with insertions, the sequence lengths of $X_t$ can change over time. To simplify notation, \citet{havasi2025edit} used an augmented space of $(X_t, Z_t)$, where it is basically always enforced that $X_t = \strip(Z_t)$. The role of $Z_t$ is only for training, to keep track of which tokens are deleted and to compute the loss, and it is neither seen by the model nor used during sampling.

To briefly summarize the construction below, the Flow Matching recipe makes use of a prescribed conditional CTMC that generates single data sequences, which is then marginalized over the data distribution. The resulting marginal CTMC will then sample from the data distribution.

\paragraph{Conditional probability path.} Given a data sequence $X_1 \sim p_\text{data}$, we prescribe a conditional probability path over $Z_t$ of the same sequence length which interpolates between the empty sequence and this data sequence. We then obtain $X_t$ by applying the $\strip$ function. Concretely, we can express the conditional probability path as
\begin{align}
    p_t(X_t, Z_t|X_1) & = p_t(X_t|Z_t, X_1) \cdot p_t(Z_t|X_1)\\
    & = p_t(X_t|Z_t) \cdot p_t(Z_t|Z_1)\\
    & = \delta_{\strip(Z_t)}(X_t) \cdot \left( \prod_{i=1}^{n} (1 - \kappa_t) \delta_{\blank}(Z_t^i) + \kappa_t \delta_{X_1^i}(Z_t^i) \right),
    \label{eq:cond_pt}
\end{align}
where $\kappa_t$ is a scheduler where $\kappa_0=0, \kappa_1=1$, and $n$ is the sequence length of $X_1$. In English, \eqref{eq:cond_pt} is a mixture distribution where each token $Z_t^i$ can either be equal to $\blank$ with probability $1 - \kappa_t$ or equal to data value $X_1^i$ with probability $\kappa_t$.

\paragraph{Conditional CTMC rate.} As discussed in \cite{havasi2025edit}, a conditional CTMC that samples from this conditional probability path can be constructed as
\begin{equation}\label{eq:cond_ut}
\begin{split}
    &u_t(x, z | X_t, Z_t, X_1) =
    \left( \sum_{i=1}^n \frac{\dot{\kappa}_t}{1 - \kappa_t} (\delta_{X_1^i}(z^i) - \delta_{Z_t^i}(z^i) ) \right) \delta_{\strip(z)}(x), \\
    & \qquad\qquad \text{ where } x = \ins(X_t, i, a) \text{ for some $i \in [n]$ and $a \in [M]$}
\end{split}
\end{equation}
which denotes the infinitesimal change in probability of going from the state $(X_t, Z_t) \rightarrow (x, z)$, constrained to next sequences $x$ that are one token insertion difference from $X_t$. In English, \eqref{eq:cond_ut} assigns a rate of $\frac{\dot{\kappa}_t}{1 - \kappa_t}$ if $Z_t^i$ is not yet equal to $X_1^i$; otherwise, it is zero. This ensures that a sample starting with all blanks $Z_0 = [\blank, ... \blank]$ at $t=0$ will eventually turn into $X_1$ at $t=1$.
This ratio $\frac{\dot{\kappa}_t}{1 - \kappa_t}$ is the infinitesimal rate that each token changes its value, matching the distribution imposed by the scheduler $\kappa_t$, and conditioned on that it is still the $\blank$ token at time $t$.

\begin{figure}
    \centering
    \includegraphics[width=0.7\linewidth]{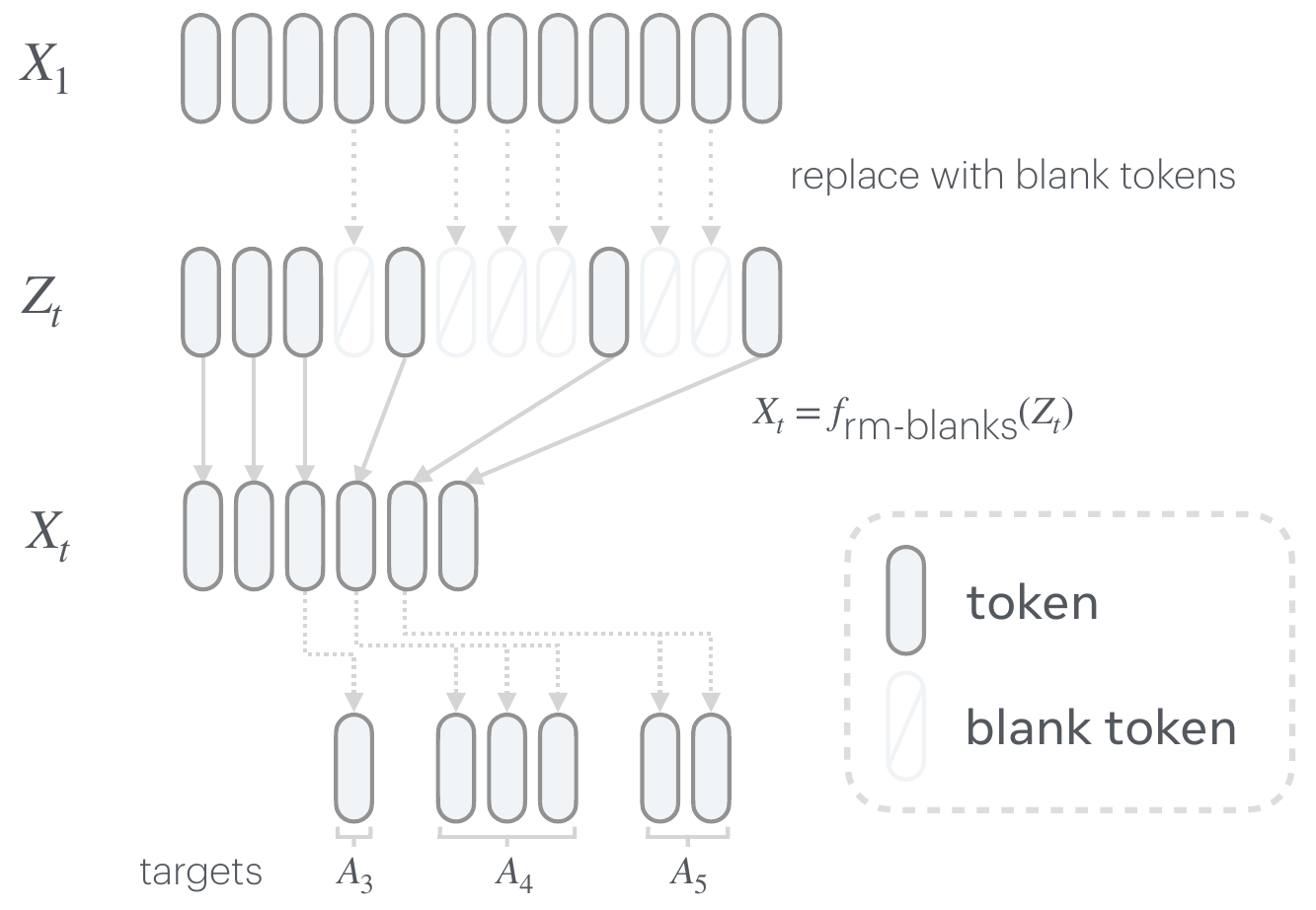}
    \caption{During training we construct $Z_t$ by replacing tokens with the blank token ($\blank$), with the original tokens used to construct the target bag-of-tokens $\mathcal{A}_i$.}
    \label{fig:training_illustration}
\end{figure}

\paragraph{Training loss.} In order to train a model thats transport sequences via insertions,
\begin{equation}
    u_t^\theta(x | X_t), \quad \text{ where } x = \ins(X_t, i, a) \text{ for some $i$ and $a$}
\end{equation}
we would need to marginalize out the auxiliary process $Z_t$ and the data $X_1$. \citet{havasi2025edit} showed this can be done by using a loss based on any Bregman divergence---following the recipe of \citet{holderrieth2024generator}---while summing up over all possible sequences $z$ such that $x = \strip(z)$. Concretely, given a convex function $\phi$ that defines a Bregman divergence $D_\phi(a, b) = \phi(a) - \phi(b) - \langle a - b, \tfrac{\mathrm{d}}{\mathrm{d} b} \phi (b) \rangle$, we can use the loss
\begin{equation}\label{eq:aux_bregman_divergence_loss}
    \E_{X_t, Z_t \sim p_t(X_t, Z_t | X_1), X_1 \sim p_\text{data}} D_\phi\Big( \tsum_{z} u_t(\cdot, z | X_t, Z_t, X_1), u_t^\theta(\cdot | X_t) \Big).
\end{equation}
Plugging in the entropy $\phi(u) = \langle u, \log u\rangle$, this results in the Edit Flow loss
\begin{equation}\label{eq:original_training_loss}
    \E_{t, p_t(X_t, Z_t | X_1), X_1 \sim p_\text{data}} \left[ \sum_{x\neq X_t} u_t^\theta(x | X_t) - \sum_{i=1}^n \1_{[Z_t^i = \blank]} \frac{\dot{\kappa}_t}{1 - \kappa_t}  \log u_t^\theta( \ins(X_t, j, X_1^i) | X_t) \right],
\end{equation}
where $j$ is the position in $X_t$ that corresponds to the first non-$\blank$ token on the left of $Z_t^i$. This ensures that inserting at the $i$-th position corresponds to changing the value of $Z_t^i$ from $\blank$ to $X_1^i$.

\paragraph{Loss simplification.} We deviate from \citet{havasi2025edit} and use a $t$-independent parameterization. In particular, for sequences $x$ that are one token insertion of $X_t$, \ie, $x = \ins(X_t, i, a)$, we use\looseness=-1
\begin{equation}
    u_t^\theta \big( \ins(X_t, i, a) | X_t \big) = \frac{\dot{\kappa}_t}{1 - \kappa_t} \lambda^i(X_t) Q^i(a | X_t),
\end{equation}
where the neural network parameterizes $\lambda$ and $Q$.
Using this parameterization, letting $\mathcal{A}_j$ be the set of missing tokens to the right of position $j$ of $X_t$, 
the training loss \eqref{eq:original_training_loss} can be decomposed into
\begin{align}
    &\E_{(\dots)} \left(\frac{\dot{\kappa}_t}{1 - \kappa_t} \right)\Bigg( 
    \sum_{j=1}^{n(X_t)} \lambda^j(X_t) - \sum_{j=1}^{n(X_t)} \sum_{a \in \mathcal{A}_j} \log \left( \lambda^j(X_t)Q^j(a | X_t) \right)
    \Bigg) \\
    =& \E_{(\dots)} \left(\frac{\dot{\kappa}_t}{1 - \kappa_t} \right)\sum_{j=1}^{n(X_t)} \Bigg( \underbrace{\vphantom{\sum_{a \in \mathcal{A}_j}} \lambda^j(X_t) - |\mathcal{A}_i| \log \lambda^j(X_t)}_{\eqref{eq:poisson_nll}} + \underbrace{\sum_{a \in \mathcal{A}_j} \log Q^j(a | X_t)}_{\eqref{eq:token_loss}}
    \Bigg) + \text{const.}
\end{align}
which recovers the losses for $\lambda$ and $Q$ in \eqref{eq:poisson_nll} and \eqref{eq:token_loss} respectively, after removing the coefficient $\frac{\dot{\kappa}_t}{1 - \kappa_t}$. While keeping this coefficient relates the loss the to an evidence lower bound \citep{havasi2025edit}, we found that removing this coefficient in the loss gave better results in practice.

\subsection{Interleaved time schedule}\label{app:interleaved_schedule}

In order to model image insertions, we would make a choice. We can either (i) fully denoise images at the time of insertion, or (ii) insert only noise and denoise later. We choose the latter approach, as this allows simultaneous generation across images and text, and provides the best parallelism as only a single model forward at each step is needed for both modalities. 

Without loss of generality, assume there is at most just one image. 
Generation starts by advancing the sequence time, denoted $t_\text{text} = 0$. When the image is inserted, we associate the image with its own time $t_\text{img}$.

The main difficulty is that we can not simply set $t_\text{img}$ and $t_\text{text}$ independently during training, as evidently we always have $t_\text{text} \geq t_\text{img}$. In fact, an independent scheduler induces the wrong distribution for our insertion prediction, and it will not insert the correct distribution at generation time. Instead, we need to ensure that training and generation see the same distribution of time values. To achieve this, we first note that the image exists in the sequence according to the scheduler $\kappa_t$, which means that the \emph{insertion times} are distributed according to
\begin{equation}
    p(t_\text{insert}) = \dot{\kappa}_t,
\end{equation}
where $t_\text{insert}$ is the time at which an image is inserted, \ie, $\kappa_t$ is the cumulative distribution function (CDF) of the insertion times. Equivalently, to sample the insertion time, we can apply the inverse CDF sampling,
\begin{equation}
    t_\text{insert} = \kappa^{-1}(u), \qquad u \sim \text{Unif}(0, 1).
\end{equation}
If we set $t_\text{img} = 0$ when an image is inserted, then the difference between $t_\text{text}$ and $t_\text{img}$ is distributed according to the insertion time. This gives us the relation
\begin{equation}
    t_\text{text} - t_\text{img} = t_\text{insert}
\end{equation}
when $0 \leq t_\text{text}, t_\text{img}, t_\text{insert} \leq 1$. Since $t_\text{text}$ will reach 1 before $t_\text{img}$, and we want to train for the entire process until $t_\text{img} = 1$, we can construct an extended time interval
\begin{equation}
    \tau_\text{text} \in [0, 2], \qquad t_\text{text} = \text{clip}(\tau_\text{text}),
\end{equation}
where $\text{clip}(\tau) = \min\{1, \max\{0, \tau\}\}$ clips the time values back into the interval $[0, 1]$.

During training, we first sample $\tau_\text{text}$, then sample
\begin{equation}
    \tau_\text{img} = \tau_{\text{text}} - \kappa^{-1}(u), \qquad u \sim \text{Unif}(0, 1).
\end{equation}
This will sample an extended time for the image in the internal [-1, 2]. If $\tau_\text{img} < 0$, then it has not yet been inserted, hence it is deleted from the sequence. Otherwise, it is clipped,
\begin{equation}
    t_\text{img} = \text{clip}(\tau_\text{img}),
\end{equation}
and we proceed to use the Flow Matching loss \eqref{eq:flow_matching} to train the image denoising.

\newpage
\section{Sampling and Training Algorithms}
\label{appendix:algorithms}

\begin{algorithm}[H]
  \caption{{\bf OneFlow interleaved text--image generation.}}
  \begin{algorithmic}[1]
    \Function{OneFlowGeneration}{step size $\Delta t$, schedule $\kappa$}
      \State $X \gets$ empty sequence \Comment{\emph{Text tokens (initially empty set)}}
      \State $\mathcal{I} \gets \emptyset$ \Comment{\emph{Set of image latents with per-image times}}
      \State $t_{\text{text}} \gets 0$
      \While{$t_{\text{text}} < 1$ \textbf{or} $\exists Y\in\mathcal{I}: t_{\text{img}}(Y) < 1$}
        \State $X, \mathcal{I}, t_{\text{text}}, t_{\text{img}} \gets \textsc{OneFlowStep}(X, \mathcal{I}, t_{\text{text}}, t_{\text{img}}, \Delta t, \kappa)$
      \EndWhile
      \State \textbf{return} $X$ and $\{\text{VAEDec}(Y): Y\in\mathcal{I}\}$ \Comment{\emph{Decode VAE latents into image space}}
    \EndFunction
  \end{algorithmic}
\end{algorithm}

\begin{algorithm}[H]
    \caption{{\bf \EF step function.} \\
    $X$ is the token sequence, $\mathcal{I}$ is the set of image latents each with time $t_{\text{img}}(Y)$.}
    \label{alg:sampling}
    \begin{algorithmic}[1]
        \Function{OneFlowStep}{$X, \mathcal{I}, t_{\text{text}}, t_{\text{img}}, \Delta t, \kappa$}
            \State $(\{\pi,\lambda_{\mathrm{nonzero}},Q\}, \{v(Y,\cdot)\}_{Y\in\mathcal{I}}) \gets \texttt{OneFlowModel}(X,\mathcal{I}, t_{\text{img}})$ 
            
            \Statex
            \ForAll{$Y \in \mathcal{I}$ \textbf{with} $t_{\text{img}}(Y) < 1$} \Comment{\emph{Image: Flow matching step on images}}
                \State $\Delta t_{\text{img}} \gets \min\{1 - t_{\text{img}}(Y), \Delta t\}$
                \State $Y \gets Y + \Delta t_{\text{img}} \cdot v(Y, t_{\text{img}}(Y))$
                \State $t_{\text{img}}(Y) \gets t_{\text{img}}(Y) + \Delta t_{\text{img}}$
            \EndFor
            \Statex
            \State $\Delta t_{\text{text}} \gets \min\{1 - t_{\text{text}}, \Delta t\}$
            \If{$\Delta t_{\text{text}} > 0$}
                \ForAll{positions $i \in \{1,\dots, n(X)\}$} \Comment{\emph{Text: parallel insertions}}
                    \State $p_i^{\pi} \gets 1-\pi^i$ \Comment{\emph{If using \eqref{eq:poisson_nll} without $\pi$, then skip this step}}
                    \State $p_i^\lambda \gets  \Delta t_{\text{text}} \cdot \frac{\dot{\kappa}(t_{\text{text}})}{1-\kappa(t_{\text{text}})} \cdot \lambda^i_{\mathrm{nonzero}}$
                    \State $\text{do-insert} \gets$ Bernoulli$(p_i^{\pi})$ and Bernoulli$(p_i^\lambda)$
                    \If{do-insert} 
                        \State $a \sim Q^i(\cdot\mid X)$
                        \State $X \gets \ins(X, i, a)$
                        \Comment{If $a=\image$, this will insert $N_\text{img}$ tokens}
                        \If{$a = \image$}
                            \State $Y \sim \mathcal{N}(0,I)$,\;\; $t_{\text{img}}(Y)\gets 0$,\;\; $\mathcal{I} \gets \mathcal{I}\cup\{Y\}$
                            
                        \EndIf
                    \EndIf
                \EndFor
            \EndIf
            \Statex
            \State $t_{\text{text}} \gets t_{\text{text}} + \Delta t_{\text{text}}$
            \State \textbf{return} $X, \mathcal{I}, t_{\text{text}}, t_{\text{img}}$
        \EndFunction
    \end{algorithmic}
\end{algorithm}

\begin{algorithm}[H]
    \caption{{\bf OneFlow training loss with interleaved schedule}}
    \label{alg:training}
    \begin{algorithmic}[1]
        \Function{OneFlowTrainingStep}{data sequence $X$, image latents $\mathcal{I}$, schedule $\kappa$}
            \State $\tau_{\text{text}} \sim \mathrm{Unif}[0,2]$
            \State $t_{\text{text}} \gets \min\{1,\tau_{\text{text}}\}$
            \State $j \gets 0$
            \State $X_t \gets []$
            \ForAll{$x^i \in X$} \Comment{\emph{Keep each ground-truth token with prob $\kappa(t_{\text{text}})$ to get noisy $X_t$}}
                \If{$r < \kappa(t_{\text{text}}) \;\text{where} \; r \sim \mathrm{Unif}(0,1)$}
                    \State $X_t \gets X_t + [x^i]$
                    \State $j \gets j+1$
                    \State $\mathcal{A}_j \gets \{\}$
                \Else
                    \State $\mathcal{A}_j \gets \mathcal{A}_j \cup \{X^i\}$  \Comment{\emph{Record the deleted tokens at each position in $\mathcal{A}_j$}}
                \EndIf
            \EndFor
            \State $\mathcal{I}_t \gets \{\}$
            \ForAll{images $Y \in \mathcal{I}$}
                \State $Y_1 \gets \text{VAEEnc}(\text{img})$
                \State $u\sim \mathrm{Unif}(0,1)$
                \State $\tau_{\text{img}}(Y) \gets \tau_{\text{text}} - \kappa^{-1}(u)$
                \If{$\tau_{\text{img}} < 0$} 
                    \State add \image\@ to the appropriate $\mathcal{A}_i$ \Comment{\emph{Image is “deleted” at this snapshot}}
                \Else
                    \State $t_{\text{img}}(Y) \gets \min\{1,\tau_{\text{img}}(Y)\}$
                    \State $Y_0 \sim \mathcal{N}(0,I)$
                    \State $Y_t \gets t_{\text{img}}(Y)Y_1 + (1-t_{\text{img}}(Y))Y_0$
                    \State $\mathcal{I}_t \gets \mathcal{I}_t \cup \{Y_t\}$
                \EndIf
            \EndFor
            \Statex
            \State $\{\pi,\lambda_{\mathrm{nonzero}},Q\} \gets \texttt{OneFlowModel}(X_t, \mathcal{I}_t)$
            \Comment{\emph{Forward pass}}
            \Statex
            \State $\mathcal{L} \gets \mathcal{L}_{\text{text}} +      \mathcal{L}_{\text{image}}$ 
            \Comment{Compute text and image losses \eqref{eq:insertion_loss} and \eqref{eq:flow_matching}}

            \Statex
            \State $\Theta \gets \texttt{optimizer\_step}(\nabla \mathcal{L}; \Theta)$ \Comment{\emph{Compute gradients and update model}}
        \EndFunction
    \end{algorithmic}
\end{algorithm}

\end{document}